\pdfoutput=1

\documentclass[11pt]{article}

\usepackage[]{ACL2023}

\usepackage{times}
\usepackage{latexsym}

\usepackage[T1]{fontenc}
\usepackage[utf8]{inputenc}

\usepackage{microtype}
\usepackage{caption}
\usepackage{enumitem}
\usepackage{tabu}
\usepackage{rotating}
\usepackage{tikz}
\usepackage{capt-of,etoolbox}
\usepackage{microtype}
\usepackage{bm}
\usepackage{soul}
\usepackage{wrapfig}
\usepackage{graphbox}
\usepackage{tcolorbox}
\usepackage{amsmath}
\usepackage{cleveref}
\usepackage{mathtools}
\usepackage{balance}
\usepackage[varqu]{zi4}
\usepackage[bb=boondox]{mathalfa}
\usepackage{units}
\usepackage{setspace}
\usepackage{gensymb}
\usepackage{helvet}
\usepackage{printlen} %
\usepackage{placeins} %
\usepackage{afterpage} %
\usepackage{csquotes}
\usepackage{hyperref}

\usepackage{eqnarray}
\usepackage{bbm}

\definecolor{redOV}{RGB}{255, 235, 238}
\definecolor{redI}{RGB}{255, 205, 210}
\definecolor{redII}{RGB}{239, 154, 154}
\definecolor{redIII}{RGB}{229, 115, 115}
\definecolor{redIV}{RGB}{239, 83, 80}
\definecolor{redV}{RGB}{244, 67, 54}
\definecolor{redVI}{RGB}{229, 57, 53}
\definecolor{redVII}{RGB}{211, 47, 47}
\definecolor{redVIII}{RGB}{198, 40, 40}
\definecolor{redIX}{RGB}{183, 28, 28}
\definecolor{redAI}{RGB}{255, 138, 128}
\definecolor{redAII}{RGB}{255, 82, 82}
\definecolor{redAIV}{RGB}{255, 23, 68}
\definecolor{redAVII}{RGB}{213, 0, 0}

\definecolor{pinkOV}{RGB}{252, 228, 236}
\definecolor{pinkI}{RGB}{248, 187, 208}
\definecolor{pinkII}{RGB}{244, 143, 177}
\definecolor{pinkIII}{RGB}{240, 98, 146}
\definecolor{pinkIV}{RGB}{236, 64, 122}
\definecolor{pinkV}{RGB}{233, 30, 99}
\definecolor{pinkVI}{RGB}{216, 27, 96}
\definecolor{pinkVII}{RGB}{194, 24, 91}
\definecolor{pinkVIII}{RGB}{173, 20, 87}
\definecolor{pinkIX}{RGB}{136, 14, 79}
\definecolor{pinkAI}{RGB}{255, 128, 171}
\definecolor{pinkAII}{RGB}{255, 64, 129}
\definecolor{pinkAIV}{RGB}{245, 0, 87}
\definecolor{pinkAVII}{RGB}{197, 17, 98}

\definecolor{purpleOV}{RGB}{243, 229, 245}
\definecolor{purpleI}{RGB}{225, 190, 231}
\definecolor{purpleII}{RGB}{206, 147, 216}
\definecolor{purpleIII}{RGB}{186, 104, 200}
\definecolor{purpleIV}{RGB}{171, 71, 188}
\definecolor{purpleV}{RGB}{156, 39, 176}
\definecolor{purpleVI}{RGB}{142, 36, 170}
\definecolor{purpleVII}{RGB}{123, 31, 162}
\definecolor{purpleVIII}{RGB}{106, 27, 154}
\definecolor{purpleIX}{RGB}{74, 20, 140}
\definecolor{purpleAI}{RGB}{234, 128, 252}
\definecolor{purpleAII}{RGB}{224, 64, 251}
\definecolor{purpleAIV}{RGB}{213, 0, 249}
\definecolor{purpleAVII}{RGB}{170, 0, 255}

\definecolor{deeppurpleOV}{RGB}{237, 231, 246}
\definecolor{deeppurpleI}{RGB}{209, 196, 233}
\definecolor{deeppurpleII}{RGB}{179, 157, 219}
\definecolor{deeppurpleIII}{RGB}{149, 117, 205}
\definecolor{deeppurpleIV}{RGB}{126, 87, 194}
\definecolor{deeppurpleV}{RGB}{103, 58, 183}
\definecolor{deeppurpleVI}{RGB}{94, 53, 177}
\definecolor{deeppurpleVII}{RGB}{81, 45, 168}
\definecolor{deeppurpleVIII}{RGB}{69, 39, 160}
\definecolor{deeppurpleIX}{RGB}{49, 27, 146}
\definecolor{deeppurpleAI}{RGB}{179, 136, 255}
\definecolor{deeppurpleAII}{RGB}{124, 77, 255}
\definecolor{deeppurpleAIV}{RGB}{101, 31, 255}
\definecolor{deeppurpleAVII}{RGB}{98, 0, 234}

\definecolor{indigoOV}{RGB}{232, 234, 246}
\definecolor{indigoI}{RGB}{197, 202, 233}
\definecolor{indigoII}{RGB}{159, 168, 218}
\definecolor{indigoIII}{RGB}{121, 134, 203}
\definecolor{indigoIV}{RGB}{92, 107, 192}
\definecolor{indigoV}{RGB}{63, 81, 181}
\definecolor{indigoVI}{RGB}{57, 73, 171}
\definecolor{indigoVII}{RGB}{48, 63, 159}
\definecolor{indigoVIII}{RGB}{40, 53, 147}
\definecolor{indigoIX}{RGB}{26, 35, 126}
\definecolor{indigoAI}{RGB}{140, 158, 255}
\definecolor{indigoAII}{RGB}{83, 109, 254}
\definecolor{indigoAIV}{RGB}{61, 90, 254}
\definecolor{indigoAVII}{RGB}{48, 79, 254}

\definecolor{blueOV}{RGB}{227, 242, 253}
\definecolor{blueI}{RGB}{187, 222, 251}
\definecolor{blueII}{RGB}{144, 202, 249}
\definecolor{blueIII}{RGB}{100, 181, 246}
\definecolor{blueIV}{RGB}{66, 165, 245}
\definecolor{blueV}{RGB}{33, 150, 243}
\definecolor{blueVI}{RGB}{30, 136, 229}
\definecolor{blueVII}{RGB}{25, 118, 210}
\definecolor{blueVIII}{RGB}{21, 101, 192}
\definecolor{blueIX}{RGB}{13, 71, 161}
\definecolor{blueAI}{RGB}{130, 177, 255}
\definecolor{blueAII}{RGB}{68, 138, 255}
\definecolor{blueAIV}{RGB}{41, 121, 255}
\definecolor{blueAVII}{RGB}{41, 98, 255}

\definecolor{lightblueOV}{RGB}{225, 245, 254}
\definecolor{lightblueI}{RGB}{179, 229, 252}
\definecolor{lightblueII}{RGB}{129, 212, 250}
\definecolor{lightblueIII}{RGB}{79, 195, 247}
\definecolor{lightblueIV}{RGB}{41, 182, 246}
\definecolor{lightblueV}{RGB}{3, 169, 244}
\definecolor{lightblueVI}{RGB}{3, 155, 229}
\definecolor{lightblueVII}{RGB}{2, 136, 209}
\definecolor{lightblueVIII}{RGB}{2, 119, 189}
\definecolor{lightblueIX}{RGB}{1, 87, 155}
\definecolor{lightblueAI}{RGB}{128, 216, 255}
\definecolor{lightblueAII}{RGB}{64, 196, 255}
\definecolor{lightblueAIV}{RGB}{0, 176, 255}
\definecolor{lightblueAVII}{RGB}{0, 145, 234}

\definecolor{cyanOV}{RGB}{224, 247, 250}
\definecolor{cyanI}{RGB}{178, 235, 242}
\definecolor{cyanII}{RGB}{128, 222, 234}
\definecolor{cyanIII}{RGB}{77, 208, 225}
\definecolor{cyanIV}{RGB}{38, 198, 218}
\definecolor{cyanV}{RGB}{0, 188, 212}
\definecolor{cyanVI}{RGB}{0, 172, 193}
\definecolor{cyanVII}{RGB}{0, 151, 167}
\definecolor{cyanVIII}{RGB}{0, 131, 143}
\definecolor{cyanIX}{RGB}{0, 96, 100}
\definecolor{cyanAI}{RGB}{132, 255, 255}
\definecolor{cyanAII}{RGB}{24, 255, 255}
\definecolor{cyanAIV}{RGB}{0, 229, 255}
\definecolor{cyanAVII}{RGB}{0, 184, 212}

\definecolor{tealOV}{RGB}{224, 242, 241}
\definecolor{tealI}{RGB}{178, 223, 219}
\definecolor{tealII}{RGB}{128, 203, 196}
\definecolor{tealIII}{RGB}{77, 182, 172}
\definecolor{tealIV}{RGB}{38, 166, 154}
\definecolor{tealV}{RGB}{0, 150, 136}
\definecolor{tealVI}{RGB}{0, 137, 123}
\definecolor{tealVII}{RGB}{0, 121, 107}
\definecolor{tealVIII}{RGB}{0, 105, 92}
\definecolor{tealIX}{RGB}{0, 77, 64}
\definecolor{tealAI}{RGB}{167, 255, 235}
\definecolor{tealAII}{RGB}{100, 255, 218}
\definecolor{tealAIV}{RGB}{29, 233, 182}
\definecolor{tealAVII}{RGB}{0, 191, 165}

\definecolor{greenOV}{RGB}{232, 245, 233}
\definecolor{greenI}{RGB}{200, 230, 201}
\definecolor{greenII}{RGB}{165, 214, 167}
\definecolor{greenIII}{RGB}{129, 199, 132}
\definecolor{greenIV}{RGB}{102, 187, 106}
\definecolor{greenV}{RGB}{76, 175, 80}
\definecolor{greenVI}{RGB}{67, 160, 71}
\definecolor{greenVII}{RGB}{56, 142, 60}
\definecolor{greenVIII}{RGB}{46, 125, 50}
\definecolor{greenIX}{RGB}{27, 94, 32}
\definecolor{greenAI}{RGB}{185, 246, 202}
\definecolor{greenAII}{RGB}{105, 240, 174}
\definecolor{greenAIV}{RGB}{0, 230, 118}
\definecolor{greenAVII}{RGB}{0, 200, 83}

\definecolor{lightgreenOV}{RGB}{241, 248, 233}
\definecolor{lightgreenI}{RGB}{220, 237, 200}
\definecolor{lightgreenII}{RGB}{197, 225, 165}
\definecolor{lightgreenIII}{RGB}{174, 213, 129}
\definecolor{lightgreenIV}{RGB}{156, 204, 101}
\definecolor{lightgreenV}{RGB}{139, 195, 74}
\definecolor{lightgreenVI}{RGB}{124, 179, 66}
\definecolor{lightgreenVII}{RGB}{104, 159, 56}
\definecolor{lightgreenVIII}{RGB}{85, 139, 47}
\definecolor{lightgreenIX}{RGB}{51, 105, 30}
\definecolor{lightgreenAI}{RGB}{204, 255, 144}
\definecolor{lightgreenAII}{RGB}{178, 255, 89}
\definecolor{lightgreenAIV}{RGB}{118, 255, 3}
\definecolor{lightgreenAVII}{RGB}{100, 221, 23}

\definecolor{limeOV}{RGB}{249, 251, 231}
\definecolor{limeI}{RGB}{240, 244, 195}
\definecolor{limeII}{RGB}{230, 238, 156}
\definecolor{limeIII}{RGB}{220, 231, 117}
\definecolor{limeIV}{RGB}{212, 225, 87}
\definecolor{limeV}{RGB}{205, 220, 57}
\definecolor{limeVI}{RGB}{192, 202, 51}
\definecolor{limeVII}{RGB}{175, 180, 43}
\definecolor{limeVIII}{RGB}{158, 157, 36}
\definecolor{limeIX}{RGB}{130, 119, 23}
\definecolor{limeAI}{RGB}{244, 255, 129}
\definecolor{limeAII}{RGB}{238, 255, 65}
\definecolor{limeAIV}{RGB}{198, 255, 0}
\definecolor{limeAVII}{RGB}{174, 234, 0}

\definecolor{yellowOV}{RGB}{255, 253, 231}
\definecolor{yellowI}{RGB}{255, 249, 196}
\definecolor{yellowII}{RGB}{255, 245, 157}
\definecolor{yellowIII}{RGB}{255, 241, 118}
\definecolor{yellowIV}{RGB}{255, 238, 88}
\definecolor{yellowV}{RGB}{255, 235, 59}
\definecolor{yellowVI}{RGB}{253, 216, 53}
\definecolor{yellowVII}{RGB}{251, 192, 45}
\definecolor{yellowVIII}{RGB}{249, 168, 37}
\definecolor{yellowIX}{RGB}{245, 127, 23}
\definecolor{yellowAI}{RGB}{255, 255, 141}
\definecolor{yellowAII}{RGB}{255, 255, 0}
\definecolor{yellowAIV}{RGB}{255, 234, 0}
\definecolor{yellowAVII}{RGB}{255, 214, 0}

\definecolor{amberOV}{RGB}{255, 248, 225}
\definecolor{amberI}{RGB}{255, 236, 179}
\definecolor{amberII}{RGB}{255, 224, 130}
\definecolor{amberIII}{RGB}{255, 213, 79}
\definecolor{amberIV}{RGB}{255, 202, 40}
\definecolor{amberV}{RGB}{255, 193, 7}
\definecolor{amberVI}{RGB}{255, 179, 0}
\definecolor{amberVII}{RGB}{255, 160, 0}
\definecolor{amberVIII}{RGB}{255, 143, 0}
\definecolor{amberIX}{RGB}{255, 111, 0}
\definecolor{amberAI}{RGB}{255, 229, 127}
\definecolor{amberAII}{RGB}{255, 215, 64}
\definecolor{amberAIV}{RGB}{255, 196, 0}
\definecolor{amberAVII}{RGB}{255, 171, 0}

\definecolor{orangeOV}{RGB}{255, 243, 224}
\definecolor{orangeI}{RGB}{255, 224, 178}
\definecolor{orangeII}{RGB}{255, 204, 128}
\definecolor{orangeIII}{RGB}{255, 183, 77}
\definecolor{orangeIV}{RGB}{255, 167, 38}
\definecolor{orangeV}{RGB}{255, 152, 0}
\definecolor{orangeVI}{RGB}{251, 140, 0}
\definecolor{orangeVII}{RGB}{245, 124, 0}
\definecolor{orangeVIII}{RGB}{239, 108, 0}
\definecolor{orangeIX}{RGB}{230, 81, 0}
\definecolor{orangeAI}{RGB}{255, 209, 128}
\definecolor{orangeAII}{RGB}{255, 171, 64}
\definecolor{orangeAIV}{RGB}{255, 145, 0}
\definecolor{orangeAVII}{RGB}{255, 109, 0}

\definecolor{deeporangeOV}{RGB}{251, 233, 231}
\definecolor{deeporangeI}{RGB}{255, 204, 188}
\definecolor{deeporangeII}{RGB}{255, 171, 145}
\definecolor{deeporangeIII}{RGB}{255, 138, 101}
\definecolor{deeporangeIV}{RGB}{255, 112, 67}
\definecolor{deeporangeV}{RGB}{255, 87, 34}
\definecolor{deeporangeVI}{RGB}{244, 81, 30}
\definecolor{deeporangeVII}{RGB}{230, 74, 25}
\definecolor{deeporangeVIII}{RGB}{216, 67, 21}
\definecolor{deeporangeIX}{RGB}{191, 54, 12}
\definecolor{deeporangeAI}{RGB}{255, 158, 128}
\definecolor{deeporangeAII}{RGB}{255, 110, 64}
\definecolor{deeporangeAIV}{RGB}{255, 61, 0}
\definecolor{deeporangeAVII}{RGB}{221, 44, 0}

\definecolor{brownOV}{RGB}{239, 235, 233}
\definecolor{brownI}{RGB}{215, 204, 200}
\definecolor{brownII}{RGB}{188, 170, 164}
\definecolor{brownIII}{RGB}{161, 136, 127}
\definecolor{brownIV}{RGB}{141, 110, 99}
\definecolor{brownV}{RGB}{121, 85, 72}
\definecolor{brownVI}{RGB}{109, 76, 65}
\definecolor{brownVII}{RGB}{93, 64, 55}
\definecolor{brownVIII}{RGB}{78, 52, 46}
\definecolor{brownIX}{RGB}{62, 39, 35}

\definecolor{grayOV}{RGB}{250, 250, 250}
\definecolor{grayI}{RGB}{245, 245, 245}
\definecolor{grayII}{RGB}{238, 238, 238}
\definecolor{grayIII}{RGB}{224, 224, 224}
\definecolor{grayIV}{RGB}{189, 189, 189}
\definecolor{grayV}{RGB}{158, 158, 158}
\definecolor{grayVI}{RGB}{117, 117, 117}
\definecolor{grayVII}{RGB}{97, 97, 97}
\definecolor{grayVIII}{RGB}{66, 66, 66}
\definecolor{grayIX}{RGB}{33, 33, 33}

\definecolor{bluegrayOV}{RGB}{236, 239, 241}
\definecolor{bluegrayI}{RGB}{207, 216, 220}
\definecolor{bluegrayII}{RGB}{176, 190, 197}
\definecolor{bluegrayIII}{RGB}{144, 164, 174}
\definecolor{bluegrayIV}{RGB}{120, 144, 156}
\definecolor{bluegrayV}{RGB}{96, 125, 139}
\definecolor{bluegrayVI}{RGB}{84, 110, 122}
\definecolor{bluegrayVII}{RGB}{69, 90, 100}
\definecolor{bluegrayVIII}{RGB}{55, 71, 79}
\definecolor{bluegrayIX}{RGB}{38, 50, 56}
\definecolor{bluegrayX}{RGB}{17, 23, 26}

\definecolor{myACMBlue}{cmyk}{1,0.1,0,0.1}
\definecolor{myACMYellow}{cmyk}{0,0.16,1,0}
\definecolor{myACMOrange}{cmyk}{0,0.42,1,0.01}
\definecolor{myACMRed}{cmyk}{0,0.90,0.86,0}
\definecolor{myACMLightBlue}{cmyk}{0.49,0.01,0,0}
\definecolor{myACMGreen}{cmyk}{0.20,0,1,0.19}
\definecolor{myACMPurple}{cmyk}{0.55,1,0,0.15}
\definecolor{myACMDarkBlue}{cmyk}{1,0.58,0,0.21}

\newcommand{\dscolor}[1]{\textcolor{blueIX}{#1}}

\crefname{figure}{fig.}{fig.}
\Crefname{figure}{Fig.}{Fig.}
\crefname{equation}{eq.}{eq.}
\Crefname{equation}{Eq.}{Eq.}
\crefname{section}{\S}{\S}
\creflabelformat{equation}{#2\textup{#1}#3}

\definecolor{darkblue}{rgb}{0, 0, 0.5}
\newcommand{\figpart}[1]{\textcolor{darkblue}{#1}}

\newcommand{\tool}{{\textsc{DiffusionDB}}}
\newcommand{\dbsize}{14 million}

\definecolor{soulorange}{RGB}{255, 212, 153}
\definecolor{soulgray}{RGB}{220, 220, 220}
\definecolor{soulgraylight}{RGB}{235, 235, 235}
\definecolor{soulred}{RGB}{252, 217, 218}
\definecolor{soulbluelight}{RGB}{208, 233, 253}
\definecolor{souldorangelight}{RGB}{254, 234, 212}
\colorlet{soulblue}{blueV!30}

\newcommand{\inlinefig}[2]{\protect\includegraphics[align=c, height=#1pt]{figs/#2}}

\definecolor{tagbordercolor}{rgb}{0.8, 0.8, 0.8}
\definecolor{tagbgcolor}{rgb}{0.9, 0.9, 0.9}

\newtcbox{\tagg}{nobeforeafter, colframe=tagbordercolor,
colback=tagbgcolor, boxrule=0.5pt, arc=1pt,
  boxsep=0pt,left=2pt,right=2pt,top=1.5pt,bottom=2pt,tcbox raise base}

\newtcbox{\featuretag}{on line,
  colframe=midgray,colback=lightgray,
  boxrule=0.5pt,arc=2pt,boxsep=0pt,left=2pt,right=1pt,top=1pt,bottom=1pt
}
\definecolor{tagbgcolor}{rgb}{1, 1, 1}

\definecolor{boxyellow}{RGB}{206, 171, 1}
\definecolor{boxgreen}{RGB}{14, 152, 136}
\definecolor{boxblue}{RGB}{77, 167, 223}

\setul{0.45ex}{0.2ex}

\title{\tool{}: A Large-scale Prompt Gallery Dataset for Text-to-Image Generative Models}
\newcommand{\authorspace}{\hspace{8pt}}

\author{
    Zijie J. Wang$^1$
    \authorspace{} Evan Montoya$^1$
    \authorspace{} David Munechika$^1$\\
    \authorspace{} \textbf{Haoyang Yang}$^1$
    \authorspace{} \textbf{Benjamin Hoover}$^{1,2}$
    \authorspace{} \textbf{Duen Horng Chau}$^1$ \\
    $^1$Georgia Tech \authorspace{} $^2$IBM Research \\
    \texttt{\{jayw|emontoya30|david.munechika|alexanderyang|bhoov|polo\}@gatech.edu}
}

\begin{document}
\maketitle
\begin{abstract}
  With recent advancements in diffusion models, users can generate high-quality images by writing text prompts in natural language.
  However, generating images with desired details requires proper prompts, and it is often unclear how a model reacts to different prompts or what the best prompts are.
  To help researchers tackle these critical challenges, we introduce \tool{}, the first large-scale text-to-image prompt dataset totaling 6.5TB, containing \dbsize{} images generated by Stable Diffusion, 1.8 million unique prompts, and hyperparameters specified by real users.
  We analyze the syntactic and semantic characteristics of prompts.
  We pinpoint specific hyperparameter values and prompt styles that can lead to model errors and present evidence of potentially harmful model usage, such as the generation of misinformation.
  The unprecedented scale and diversity of this human-actuated dataset provide exciting research opportunities in understanding the interplay between prompts and generative models, detecting deepfakes, and designing human-AI interaction tools to help users more easily use these models.
  \tool{} is publicly available at: \textbf{\url{https://poloclub.github.io/diffusiondb}}.\looseness=-1
\end{abstract}

\section{Introduction}

Recent diffusion models have gained immense popularity by enabling high-quality and controllable image generation based on text prompts written in natural language~\cite{rombachHighresolutionImageSynthesis2022, rameshHierarchicalTextConditionalImage2022, sahariaPhotorealisticTexttoImageDiffusion2022}.
Since the release of these models, people from different domains have quickly applied them to create award-winning artworks~\cite{rooseGeneratedPictureWon2022}, synthetic radiology images~\cite{chambonAdaptingPretrainedVisionLanguage2022}, and even hyper-realistic videos~\cite{hoImagenVideoHigh2022}.

\setlength{\belowcaptionskip}{-9pt}
\setlength{\abovecaptionskip}{5pt}
\begin{figure}[t]
    \includegraphics[width=\linewidth]{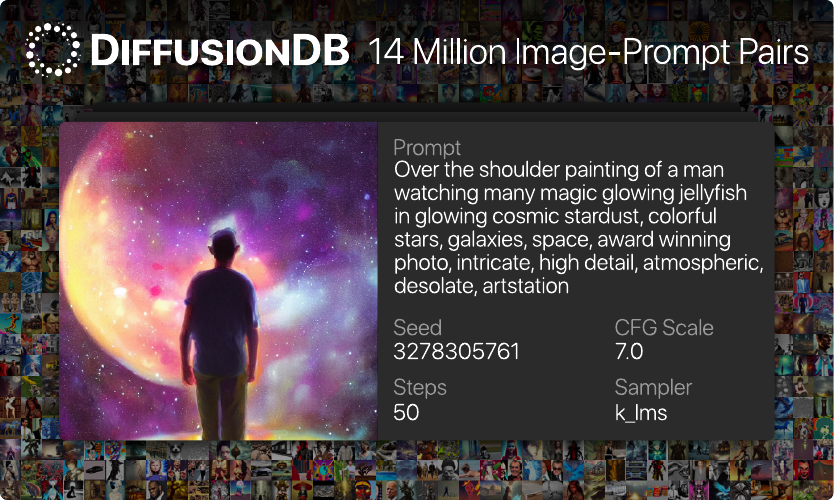}
    \centering
    \caption{
        \tool{} is the first large-scale dataset featuring 6.5TB data including 1.8 million unique Stable Diffusion prompts and 14 million generated images with accompanying hyperparameters.
        It provides exciting research opportunities in prompt engineering, deepfake detection, and understanding large generative models.
    }
    \label{fig:teaser}
\end{figure}
\setlength{\belowcaptionskip}{0pt}
\setlength{\abovecaptionskip}{10pt}

However, generating images with desired details is difficult, as it requires users to write proper prompts specifying the exact expected results.
Developing such prompts requires trial and error, and can often feel random and unprincipled~\cite{liuDesignGuidelinesPrompt2022}.
\citet{willisonStableDiffusionBreaks2022} analogize writing prompts to wizards learning ``magical spells'': users do not understand why some prompts work, but they will add these prompts to their ``spell book.''
For example, to generate highly-detailed images, it has become a common practice to add special keywords such as ``\texttt{trending on artstation}'' and ``\texttt{unreal engine}'' in the prompt.

\setlength{\belowcaptionskip}{-2pt}
\setlength{\abovecaptionskip}{9pt}
\begin{figure*}[tb]
    \includegraphics[width=\linewidth]{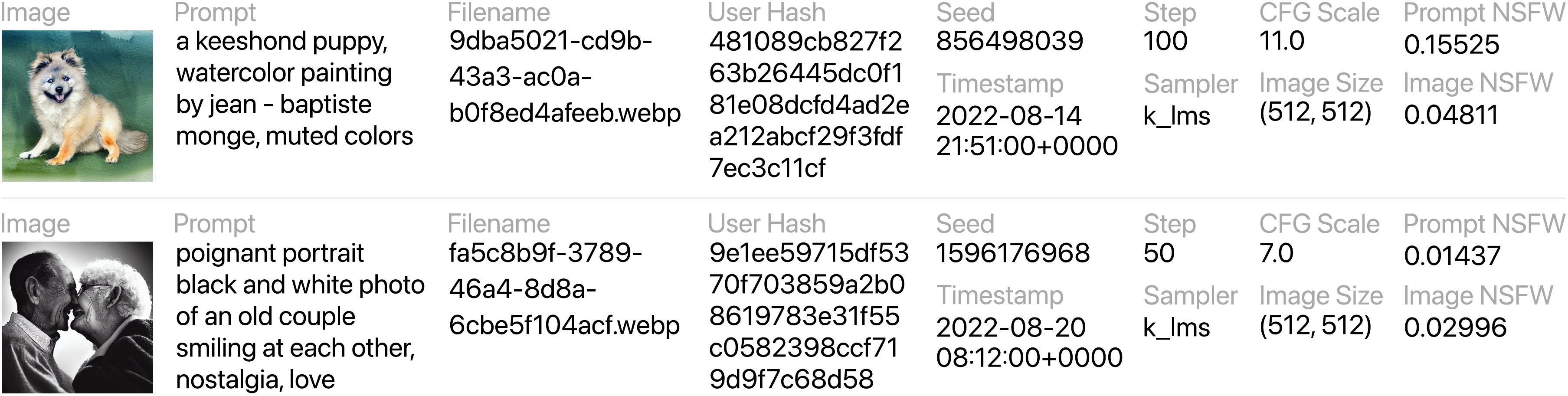}
    \centering
    \caption{
      \tool{} contains \dbsize{} Stable Diffusion images, 1.8 million unique text prompts, and all model hyperparameters: \texttt{seed}, \texttt{step}, \texttt{CFG scale}, \texttt{sampler}, and \texttt{image size}.
      Each image also has a unique filename, a hash of its creator's Discord username, and a creation timestamp.
      To help researchers filter out potentially unsafe or harmful content, we employ state-of-the-art models to compute an NSFW score for each image and prompt.\looseness=-1
    }
    \label{fig:table-view}
  \end{figure*}
\setlength{\belowcaptionskip}{0pt}
\setlength{\abovecaptionskip}{10pt}

Prompt engineering has become a field of study in the context of text-to-text generation, where researchers systematically investigate how to construct prompts to effectively solve different downstream tasks~\cite{branwenGPT3CreativeFiction2020,reynoldsPromptProgrammingLarge2021}.
As large text-to-image models are relatively new, there is a pressing need to understand how these models react to prompts, how to write effective prompts, and how to design tools to help users generate images~\cite {liuDesignGuidelinesPrompt2022}.
Our work helps researchers tackle these critical challenges, through three major \textbf{contributions}:

\begin{itemize}[topsep=3pt, itemsep=2pt, parsep=0pt, leftmargin=10pt]
    \item \textbf{\tool{}~(\autoref{fig:teaser}), the first large-scale prompt dataset totaling 6.5TB}, containing \dbsize{} images generated by Stable Diffusion~\cite{rombachHighresolutionImageSynthesis2022} using 1.8 million unique prompts and hyperparameters specified by real users.
    We construct this dataset by collecting images shared on the Stable Diffusion public Discord server~(\autoref{sec:construction}).
    We release \tool{} with a CC0 1.0 license, allowing users to flexibly share and adapt the dataset for their use.
    In addition, we open-source our code\footnote{Code: \textbf{\url{https://github.com/poloclub/diffusiondb}}} that collects, processes, and analyzes the images and prompts.
    \looseness=-1

    \item \textbf{Revealing prompt patterns and model errors.}
    The unprecedented scale of \tool{} paves the path for researchers to systematically investigate diverse prompts and associated images that were previously not possible.
    By characterizing prompts and images, we discover common prompt patterns and find different distributions of the semantic representations of prompts and images.
    Our error analysis highlights particular hyperparameters and prompt styles can lead to model errors.
    Finally, we provide evidence of image generative models being used for potentially harmful purposes such as generating misinformation and nonconsensual pornography~(\autoref{sec:analysis}).

    \item \textbf{Highlighting new research directions.}
    As the first-of-its-kind text-to-image prompt dataset, \tool{} opens up unique opportunities for researchers from natural language processing (NLP), computer vision, and human-computer interaction (HCI) communities.
    The scale and diversity of this human-actuated dataset will provide new research opportunities in better tooling for prompt engineering, explaining large generative models, and detecting deepfakes~(\autoref{sec:future}).
\end{itemize}

\noindent We believe \tool{} will serve as an important resource for researchers to study the roles of prompts in text-to-image generation and design next-generation human-AI interaction tools.

\section{Constructing \tool{}}
\label{sec:construction}

We construct \tool{}~(\autoref{fig:table-view}) by scraping user-generated images from the official Stable Diffusion Discord server.
We choose Stable Diffusion as it is currently the only open-source large text-to-image generative model, and all generated images have a CC0 1.0 license that allows uses for any purpose~\cite{stabilityaiStableDiffusionDream2022}.
We choose the official public Discord server as it has strict rules against generating illegal, hateful, or NSFW (not suitable for work, such as sexual and violent content) images, and it prohibits sharing prompts with personal information~\cite{stabilityaiStableDiffusionDiscord2022}.

Our construction process includes collecting images~(\autoref{sec:construction:collection}), linking them to prompts and hyperparameters~(\autoref{sec:construction:process}), applying NSFW detectors~(\autoref{sec:construction:nsfw}), creating a flexible file structure~(\autoref{sec:construction:organize}), and distributing the dataset~(\autoref{sec:construction:distribution}).
We discuss \tool{}'s limitations and broader impacts in \autoref{sec:limitations}, \autoref{sec:ethics}, and a Data Sheet~\cite{gebruDatasheetsDatasets2020}~(\autoref{sec:datasheet}).

\subsection{Collecting User Generated Images}
\label{sec:construction:collection}

We download chat messages from the Stable Diffusion Discord channels with DiscordChatExporter~\cite{holubDiscordChatExporterExportsDiscord2017}, saving them as HTML files.
We focus on channels where users can command a bot to run Stable Diffusion Version 1 to generate images by typing a prompt, hyperparameters, and the number of images.
The bot then replies with the generated images and used random seeds.

\subsection{Extracting Image Metadata}
\label{sec:construction:process}

We use Beautiful Soup~\cite{richardsonBeautifulSoupDocumentation2007} to parse HTML files, mapping generated images with their prompts, hyperparameters, seeds, timestamps, and the requester's Discord usernames.
Some images are collages, where the bot combines $n$ generated images as a grid (e.g., a $3\times3$ grid of $n = 9$ images);
these images have the same prompt and hyperparameters but different seeds.
We use Pillow~\cite{clarkPillowPythonImaging2015} to split a collage into $n$ individual images and assign them with the correct metadata and unique filenames.
Finally, we compress all images in \tool{} using lossless WebP~\cite{googleComparativeStudyWebP2010}.

\subsection{Identifying NSFW Content}
\label{sec:construction:nsfw}

The Stable Diffusion Discord server prohibits generating NSFW images~\cite{stabilityaiStableDiffusionDiscord2022}.
Also, Stable Diffusion has a built-in NSFW filter that automatically blurs generated images if it detects NSFW content.
However, we find \tool{} still includes NSFW images that were not detected by the built-in filter or removed by server moderators.
To help researchers filter these images, we apply state-of-the-art NSFW classifiers to compute NSFW scores for each prompt and image.
Researchers can determine a suitable threshold to filter out potentially unsafe data for their tasks.

\begin{figure}[tb]
  \includegraphics[width=\linewidth]{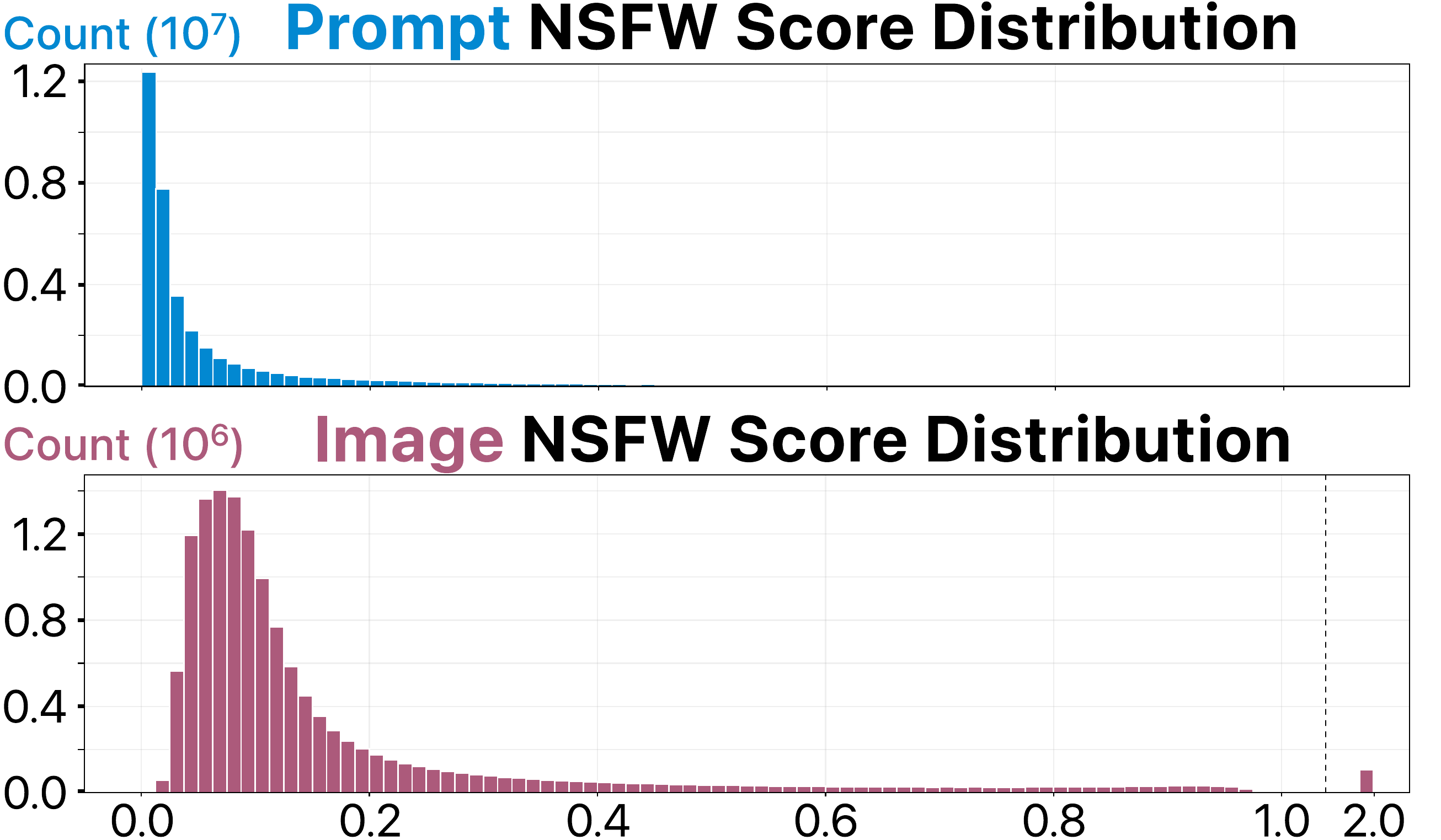}
  \centering
  \caption{
    To help researchers filter out potentially unsafe data in \tool{}, we apply NSFW detectors to predict the probability that an image-prompt pair contains NSFW content.
    For images, a score of \texttt{2.0} indicates the image has been blurred by Stable Diffusion.
  }
  \label{fig:nsfw}
\end{figure}

\paragraph{NSFW Prompts.}
We use a pre-trained multilingual toxicity prediction model to detect unsafe prompts~\cite{hanuDetoxifyToxicComment2020}.
This model outputs the probabilities of a sentence being toxic, obscene, threat, insult, identity attack, and sexually explicit.
We compute the text NSFW score by taking the maximum of the probabilities of being toxic and sexually explicit~(\autoref{fig:nsfw}~\figpart{Top}).

\paragraph{NSFW Images.}
We use a pre-trained EfficientNet classifier to detect images with sexual content~\cite{schuhmannLAION5BOpenLargescale2022}.
This model predicts the probabilities of five image types: drawing, hentai, neutral, sexual, or porn.
We compute the image NSFW score by summing the probabilities of hentai, sexual, and porn.
We use a Laplacian convolution kernel with a threshold of \texttt{10} to detect images that have already been blurred by Stable Diffusion and assign them a score of \texttt{2.0}~(\autoref{fig:nsfw}~\figpart{Bottom}).
As Stable Diffusion's blur effect is strong, our blurred image detector has high precision and recall (both 100\% on 50k randomly sampled images).
\setlength{\belowcaptionskip}{0pt}
\setlength{\abovecaptionskip}{10pt}

\paragraph{NSFW Detector Accuracy.}
To access the accuracy of these two pre-trained state-of-the-art NSFW detectors, we randomly sample 5k images and 2k prompt texts and manually annotate them with two binary NSFW labels (one for image and one for prompt) and analyze the results.
As the percentage of samples predicted as NSFW (score > 0.5) is small, we up-sample positive samples for annotation, where we have an equal number of positive and negative examples in our annotation sample.
After annotation, we compute the precisions and recalls.
Because we have up-sampled positive predictions, we adjust the recalls by multiplying false negatives by a scalar to adjust the sampling bias.
The up-sampling does not affect precisions.
Finally, the precisions, recalls and adjusted recalls are \texttt{0.3604}, \texttt{0.9565}, and \texttt{0.6661} for the prompt NSFW detector, and \texttt{0.315}, \texttt{0.9722}, and \texttt{0.3037} for the image NSFW detector.
Our results suggest two detectors are progressive classifiers.
The lower adjusted recall of the prompt NSFW detector can be attributed to several potential factors, including the use of a fixed binary threshold and the potential discrepancy in the definition of NSFW prompts between the detector and our annotation process.

\subsection{Organizing \tool{}}
\label{sec:construction:organize}

We organize \tool{} using a flexible file structure.
We first give each image a unique filename using Universally Unique Identifier (UUID, Version 4)~\cite{leachUniversallyUniqueIDentifier2005}.
Then, we organize images into 14,000 sub-folders---each includes 1,000 images.
Each sub-folder also includes a JSON file that contains 1,000 key-value pairs mapping an image name to its metadata.
An example of this image-prompt pair can be seen in~\autoref{fig:table-view}.
This modular file structure enables researchers to flexibly use a subset of \tool{}.

We create a metadata table in Apache Parquet format~\cite{apacheApacheParquetOpen2013} with 13 columns: \texttt{unique image name}, \texttt{image path}, \texttt{prompt}, \texttt{seed}, \texttt{CFG scale}, \texttt{sampler}, \texttt{width}, \texttt{height}, \texttt{username hash}, \texttt{timestamp}, \texttt{image NSFW score}, and \texttt{prompt NSFW score}.
We store the table in a column-based format for efficient querying of individual columns.

\subsection{Distributing \tool{}}
\label{sec:construction:distribution}

We distribute \tool{} by bundling each image sub-folder as a Zip file.
We collect Discord usernames of image creators~(\autoref{sec:construction:process}), but only include their SHA256 hashes in the distribution---as some prompts may include sensitive information, and explicitly linking them to their creators can cause harm.
We host our dataset on a publicly accessible repository\footnote{Public dataset repository: \textbf{\url{https://huggingface.co/datasets/poloclub/diffusiondb}}} under a CC0 1.0 license.
We provide scripts that allow users to download and load \tool{} by writing two lines of  code.
We discuss the broader impacts of our distribution in \autoref{sec:limitations}, \autoref{sec:ethics}, and the Data Sheet~(\autoref{sec:datasheet}).
To mitigate the potential harms, we provide a form for people to report harmful content for removal.
Image creators can also use this form to remove their images.

\section{Data Analysis}
\label{sec:analysis}

To gain a comprehensive understanding of the dataset, we analyze it from different perspectives.
We examine prompt length~(\autoref{sec:analysis:prompt:length}), language~(\autoref{sec:analysis:prompt:language}), characteristics of both prompts~(\autoref{sec:analysis:prompt:characteristics}) and images~(\autoref{sec:analysis:image:characteristics}).
We conduct an error analysis on misaligned prompt-image pairs~(\autoref{sec:analysis:error}) and provide empirical evidence of potentially harmful uses of image generative models~(\autoref{sec:analysis:harmful}).\looseness=-1

\setlength{\belowcaptionskip}{-10pt}
\setlength{\abovecaptionskip}{3pt}
\begin{figure}[tb]
  \includegraphics[width=\linewidth]{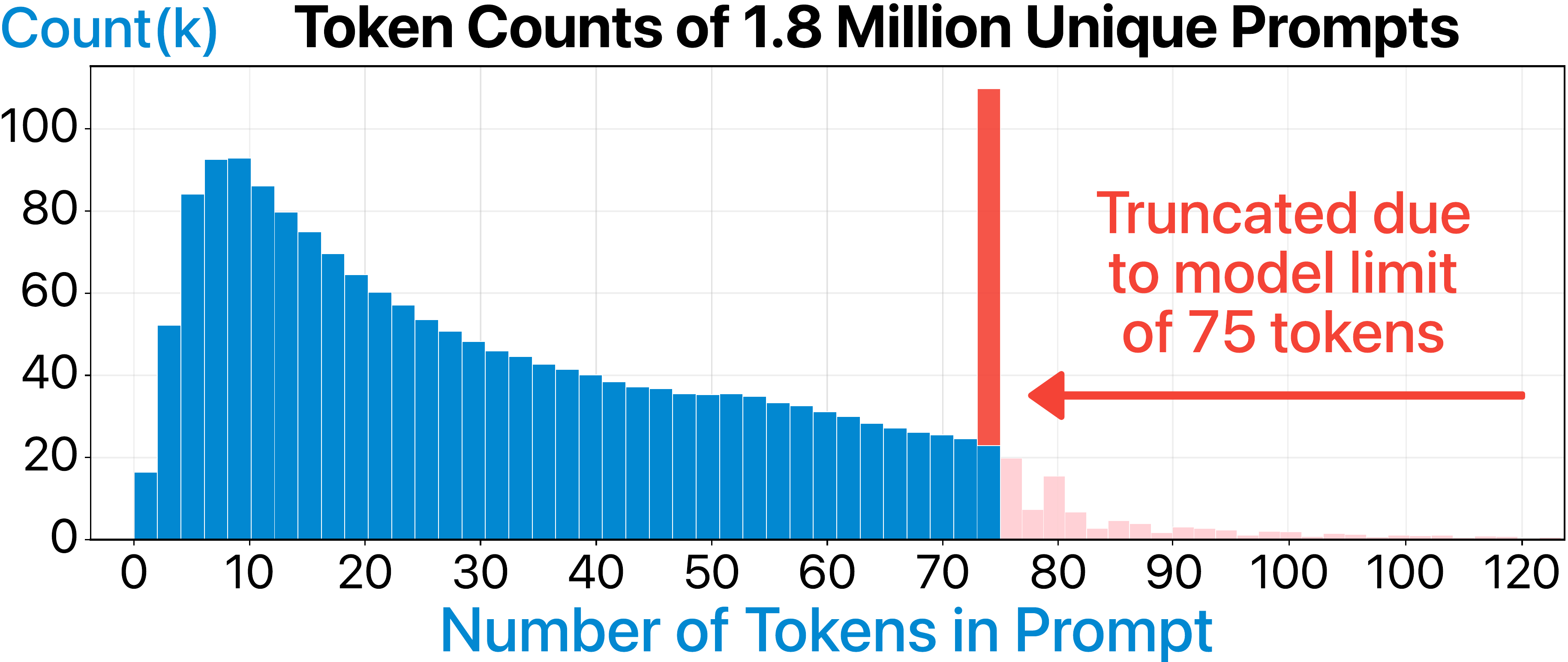}
  \centering
  \caption{
    The distribution of token counts for all 1.8 million unique prompts in \tool{}.
    It is worth noting that Stable Diffusion truncates prompts at 75 tokens.
  }
  \label{fig:tok-length}
\end{figure}
\setlength{\belowcaptionskip}{0pt}
\setlength{\abovecaptionskip}{10pt}

\begin{figure*}[tb]
  \includegraphics[width=\linewidth]{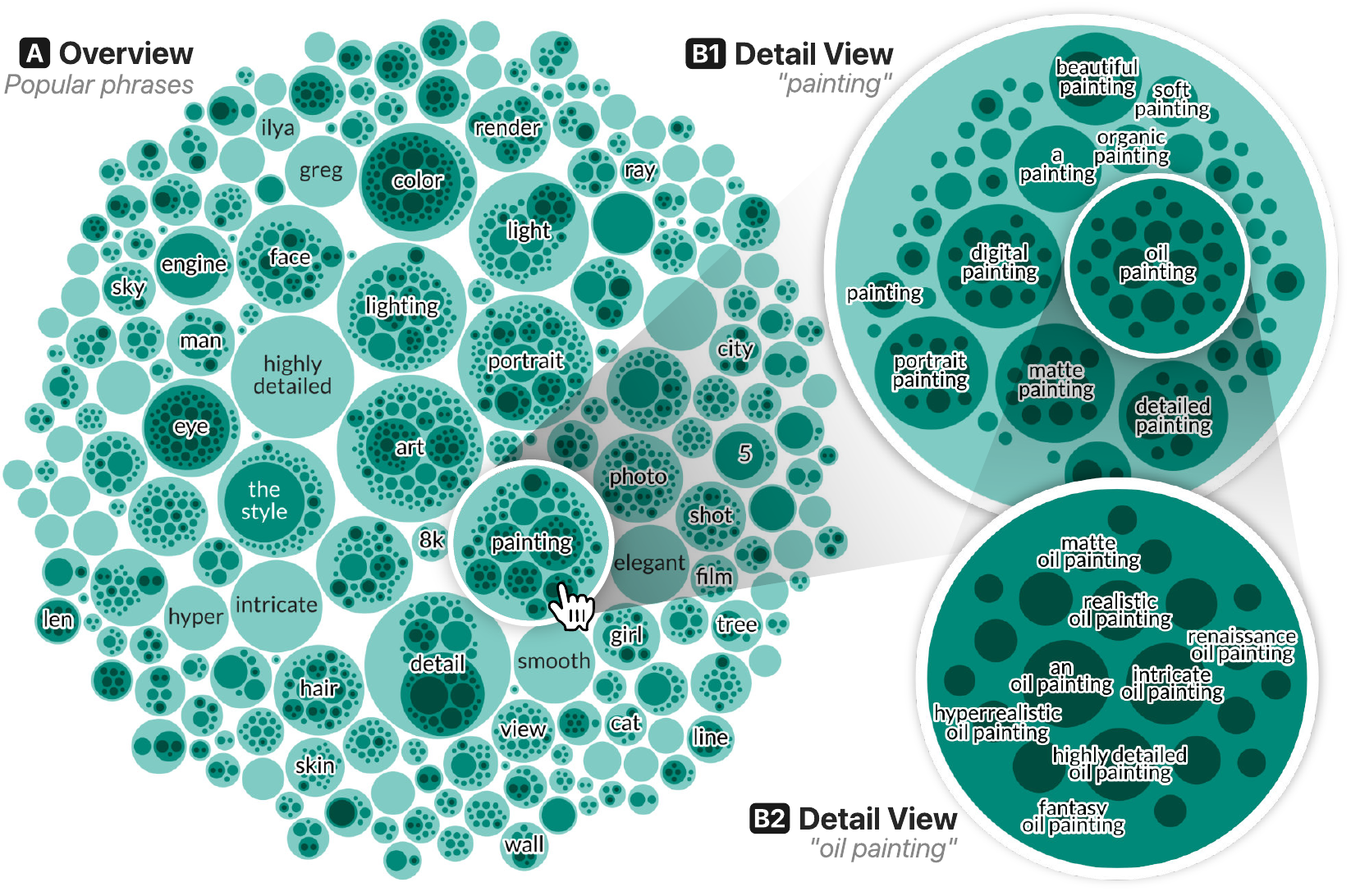}
  \centering
  \caption{
    We identify and group popular phrases in prompts through named entity recognition and dependency parsing.
    Our interactive circle-packing visualization highlights the distribution and hierarchy of these phrases.
    \textbf{(A)} The \textit{Overview} visualizes each phrase as a circle, with its size representing the phrase's frequency.
    In this example, a viewer clicks a circle to zoom into the ``painting'' phrase.
    \textbf{(B1)} The \textit{Detail View} shows all noun phrases that use ``painting'' as their root.
    \textbf{(B2)} Similarly, it shows all phrases that include ``oil painting'' when the viewer zooms into ``oil painting.''\looseness=-1
  }
  \label{fig:packing}
\end{figure*}

\subsection{Prompt Length}
\label{sec:analysis:prompt:length}

We collect prompts from Discord, where users can submit one prompt to generate multiple images and experiment with different hyperparameters.
Our dataset contains $1,819,808$ unique prompts.
We tokenize prompts using the same tokenizer as used in Stable Diffusion~\citep{platenDiffusersStateoftheartDiffusion2022}.
This tokenizer truncates tokenized prompts at $75$ tokens, excluding special tokens \texttt{<|startoftext|>} and \texttt{<|endoftext|>}.
We measure the length of prompts by their tokenized length.
The prompt length distribution~(\autoref{fig:tok-length}) indicates that shorter prompts (e.g., around 6 to 12 tokens) are the most popular.
The spike at 75 suggests many users submitted prompts longer than the model's limit, highlighting the need for user interfaces guiding users to write prompts within the token limit.

\subsection{Prompt Language}
\label{sec:analysis:prompt:language}

We use a pre-trained language detector~\cite{joulinBagTricksEfficient2017} to identify the languages used in prompts.
98.3\% of the unique prompts in our dataset are written in English.
However, we also find a large number of non-English languages, with the top four being German (5.2k unique prompts), French (4.6k), Italian (3.2k), and Spanish (3k).
The language detector identifies 34 languages with at least 100 unique prompts in total.
Stable Diffusion is trained on \texttt{LAION-2B(en)}~\cite{schuhmannLAION5BOpenLargescale2022} that primarily includes images with English descriptions, thus our findings suggest that expanding the training data's language coverage to improve the user experience for non-English communities.

\subsection{Characterizing Prompts}
\label{sec:analysis:prompt:characteristics}

In this section, we explore the characteristics of prompts in \tool{}.
We examine the syntactic~(\autoref{sec:analysis:prompt:syntactic}) and semantic~(\autoref{sec:analysis:prompt:semantic}) features of prompt text via interactive data visualizations.
Lastly, We discuss the implications of our findings and suggest future research directions.

\subsubsection{Prompt Syntactic Features}
\label{sec:analysis:prompt:syntactic}

To characterize the composition of prompts, we parse \texttt{phrases} from all 1.8M unique prompts.
We split each prompt by commas and then extract \textcolor{orangeVII}{named entities (NE)} and \textcolor{lightblueVIII}{noun phrases (NP)} from each separated component using use Spacy~\cite{honnibalSpaCyIndustrialstrengthNatural2020}.
If there is no noun phrase in a comma-separated component, we extract the \textcolor{greenVIII}{whole component (C)} as a phrase.
We keep track of each \textcolor{lightblueVIII}{NP's root} to create a hierarchy of \textcolor{lightblueVIII}{noun phrases}.

For example, for the prompt ``\texttt{draw baby yoda in a loading screen for grand theft auto 5, highly detailed, digital art, concept art},'' we extract six phrases: \textcolor{orangeVII}{``\texttt{baby yoda}'' (NE)}, \textcolor{lightblueVIII}{``\texttt{a loading screen}'' (NP with root ``screen'')}, \textcolor{orangeVII}{``\texttt{grand theft auto 5}'' (NE)}, \textcolor{greenVIII}{``\texttt{highly detailed}'' (C)}, \textcolor{lightblueVIII}{``\texttt{digital art}' (NP with root ``art'')}, and \textcolor{lightblueVIII}{``\texttt{concept art}'' (NP with root ``\texttt{art}'')}.
We group \textcolor{lightblueVIII}{``\texttt{digital art}''} and \textcolor{lightblueVIII}{``\texttt{concept art}''} into the same hierarchy as they share the same \textcolor{lightblueVIII}{NP root ``\texttt{art}.''}

\paragraph{Visualizing Prompt Phrases.}
We create an interactive circle packing visualization\footnote{Phrase visualization: \textbf{\url{https://poloclub.github.io/diffusiondb/explorer\#phrase}}} to gain an understanding of the distribution and relationships between different phrases~(\autoref{fig:packing}).
Circle packing~\cite{wangVisualizationLargeHierarchical2006} is a technique to visualize hierarchical data, and each phrase is represented as a circle whose size encodes the phrase's frequency in the dataset.
We position sibling noun phrases (e.g., phrases sharing the same NP root) inside their parent phrase's circle through a front-chain packing algorithm~\cite{wangVisualizationLargeHierarchical2006}.
Viewers can hover over a circle to see the corresponding phrase and its frequency.
Viewers can also click a circle~(\autoref{fig:packing}\figpart{A}) to zoom into that sub-tree to see more details about a phrase~(\autoref{fig:packing}\figpart{-B1}) or a sub-phrase~(\autoref{fig:packing}\figpart{-B2}).

\paragraph{Insights and implications.}
Our interactive visualization reveals that key phrases such as ``\texttt{highly detailed},'' ``\texttt{intricate},'' and ``\texttt{greg rutkowski}'' are commonly used in prompts~(\autoref{fig:packing}\figpart{A}).
The hierarchical visualization also surfaces popular image styles specified by users, such``\texttt{digital painting},'' ``\texttt{oil painting},'' and ``\texttt{portrait painting}'' for painting styles~(\autoref{fig:packing}\figpart{-B1}) and ``\texttt{studio lighting},'' ``\texttt{volumetric lighting}'', and ``\texttt{atmospheric lighting}'' for lighting.
These phrases can be unfamiliar to Stable Diffusion users, especially beginners, which highlights the importance of helping users develop prompting vocabularies.
Researchers can leverage \tool{} and our visualization to design tutorials and user interfaces that integrate exemplar prompts to guide users in describing their desired images.

\subsubsection{Prompt Semantic Features}
\label{sec:analysis:prompt:semantic}

In addition to analyzing the syntactic characteristics of prompts, we also analyze their semantic features.
We use a pre-trained CLIP model~\cite{radfordLearningTransferableVisual2021} to extract semantic features~\cite{rameshHierarchicalTextConditionalImage2022}.
We use a frozen CLIP ViT-L/14 text encoder (the same model used in Stable Diffusion) to convert prompts into 768-dimension vectors.

\paragraph{Visualizing Prompt Embeddings.}
\label{sec:analysis:prompt:visualization}

To study the distribution of prompts in high-dimensional space, we use UMAP~\cite{mcinnesUMAPUniformManifold2020} to project 768-dimensional vectors into 2-D vectors for easy visualization.
UMAP is a popular dimensionality reduction technique that is better at preserving the global structure of data and more scalable to large datasets compared to t-SNE~\cite{vandermaatenVisualizingDataUsing2008} and PCA~\cite{hotellingRelationsTwoSets1936}.
We use grid search to fine-tune hyperparameters \texttt{n\_neighbors} (\texttt{60}) and \texttt{min\_dist} (\texttt{0.1}) so that prompts are more spread out in a 2-D space.

\begin{figure}[tb]
  \includegraphics[width=\linewidth]{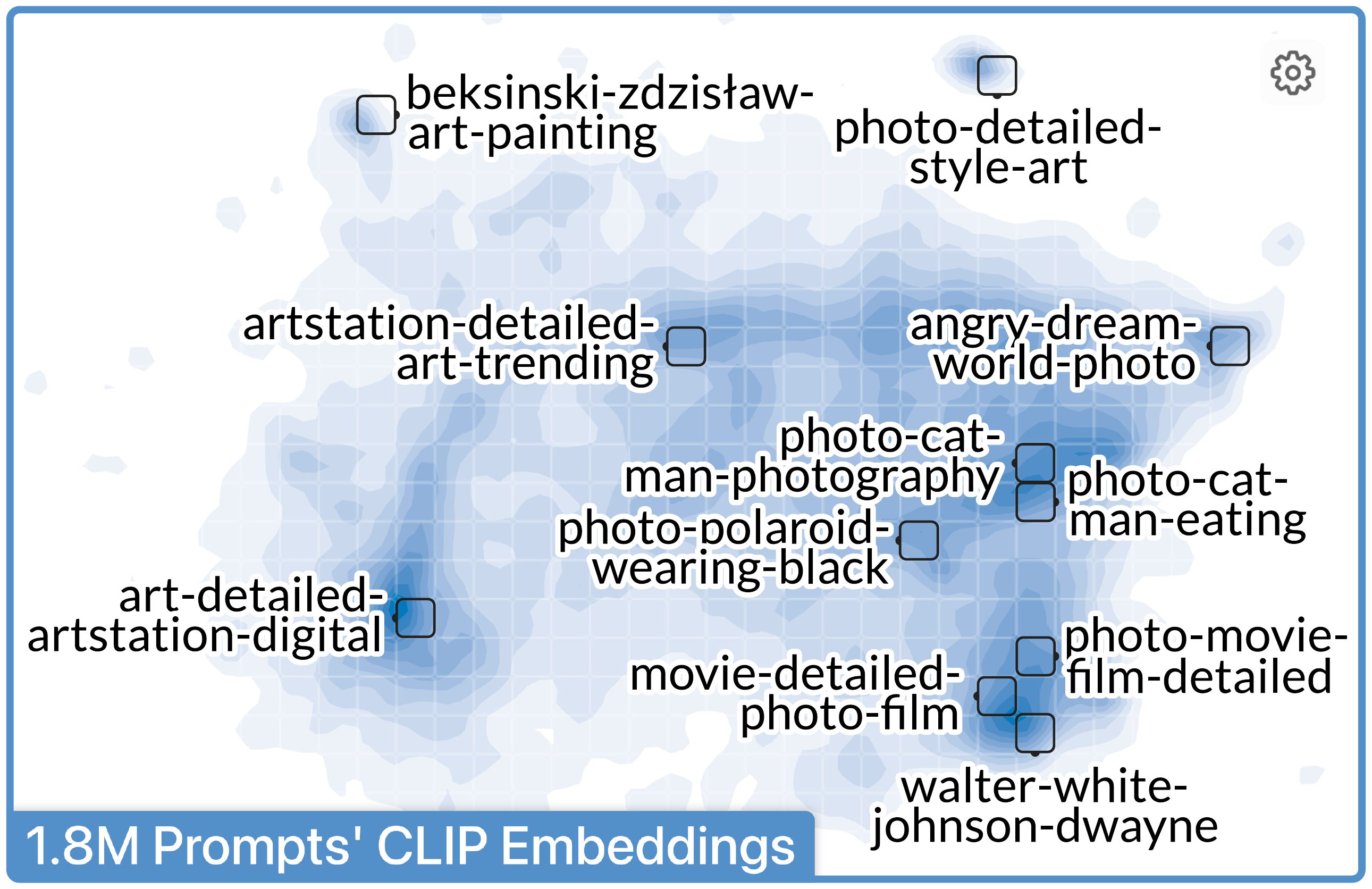}
  \centering
  \caption{
    An interactive contour plot of 1.8M prompts' CLIP embeddings, created with UMAP and kernel density estimation.
    Text labels show the top keywords of prompts in a grid tile.
    It reveals popular prompt topics.
  }
  \label{fig:text-embedding}
\end{figure}

We develop an interactive visualization tool\footnote{Prompt embedding visualization: \textbf{\url{https://poloclub.github.io/diffusiondb/explorer/\#prompt-embedding}}} to explore prompts' semantic embeddings~(\autoref{fig:text-embedding}).
We use Kernel Density Estimation (KDE)~\cite{rosenblattRemarksNonparametricEstimates1956} with a standard multivariate Gaussian kernel and Silverman bandwidth~\cite{silvermanDensityEstimationStatistics2018} to estimate the distribution of prompts' UMAP representations.
Then, we visualize the estimated distribution as a contour plot.
To summarize prompts that are in the same region, we create four grids with varying granularity and pre-compute keywords for each grid tile, by treating all prompts in the tile as a document and selecting the top 4 keywords with the highest TF-IDF scores.

\paragraph{Interactions.}
Our visualization shows keywords of tiles that are close to high-density regions and prompt clusters by default.
Viewers can hover over a tile to see its keywords, pan and zoom in to see more details of specific regions, and click a button to display each prompt as a small dot that viewers can hover over to read its prompt text.

\paragraph{Insights and implications.}
Our semantic embedding visualization~(\autoref{fig:text-embedding}) highlights two popular prompt categories: art-related prompts (left in the plot) and photography-related prompts (dark blue regions on the right).
These two groups appear distant from each other in the UMAP space, suggesting that the prompts for art and photography typically have distinct semantic representations.
Interestingly, photography prompts appear to contain two clusters: one for non-human objects (top right) and another for celebrities (bottom right).
Small prompt clusters outside the central area often feature artist names.
Our findings suggest that future researchers can leverage the prompt usage distribution to fine-tune generative models to tailor to specific popular prompt categories.

\begin{figure}[tb]
  \includegraphics[width=\linewidth]{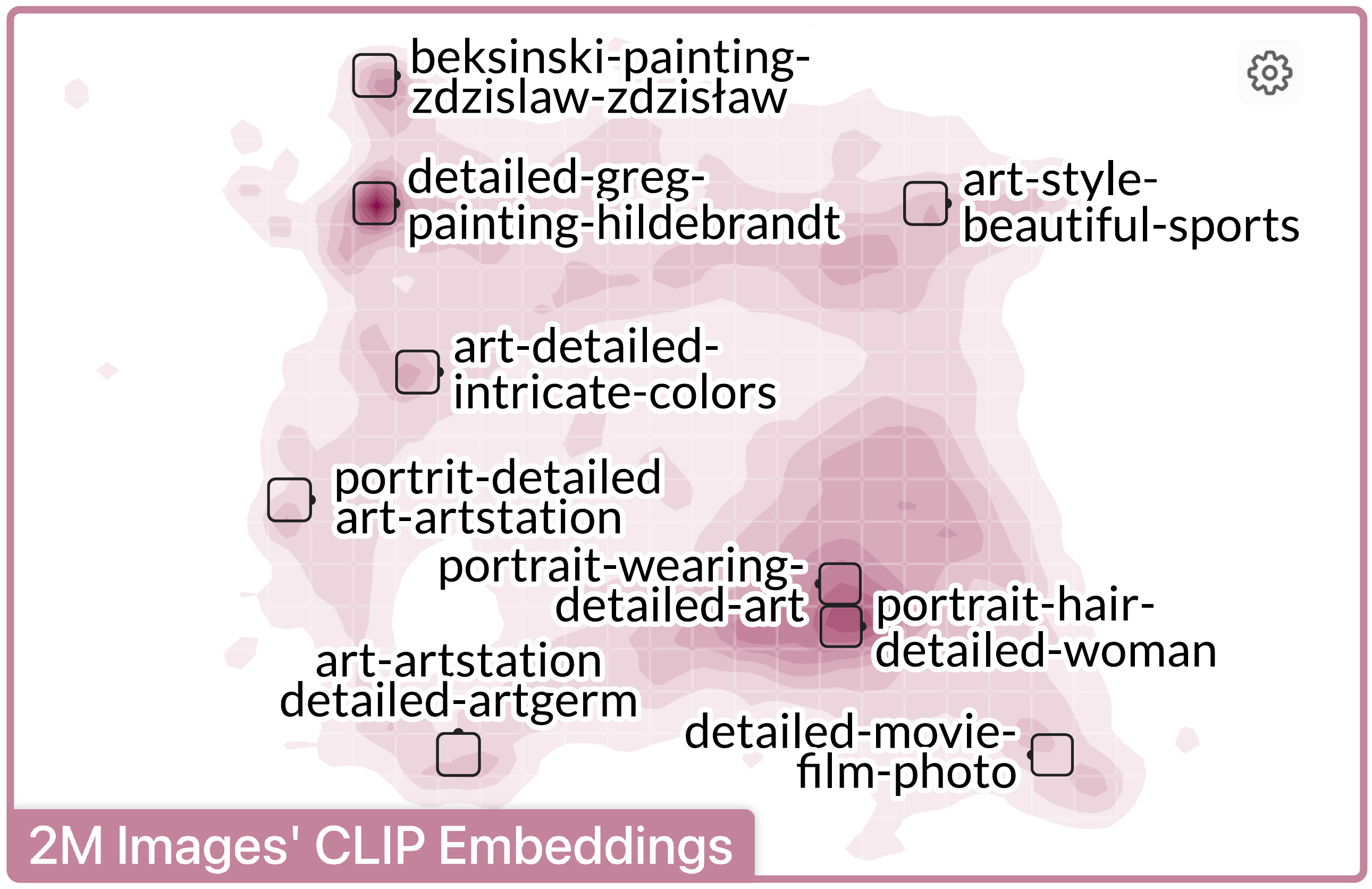}
  \centering
  \caption{
    CLIP embeddings of 2M randomly selected images shown as a contour plot, with text labels being keywords of prompts in the grid tiles.
    It shows images have a different embedding distribution from prompts.
  }
  \label{fig:image-embedding}
\end{figure}

\subsection{Characterizing Images}
\label{sec:analysis:image:characteristics}

We visualize\footnote{Image embedding visualization: \textbf{\url{https://poloclub.github.io/diffusiondb/explorer/\#image-embedding}}} the CLIP embedding distribution of 2 million unique image instances randomly sampled from \tool{}~(\autoref{fig:image-embedding}) by defining the unique key as the combination of the image's prompt and hyperparameters \texttt{CFG scale}, \texttt{step}, \texttt{size}, and \texttt{seed}.
We use the UMAP model that was previously trained on the prompt embeddings to project the image embeddings into the same 2-D space.
Finally, we apply the same method we used for our prompt embedding visualization~(\autoref{sec:analysis:prompt:visualization}) to generate a contour plot and grid label overlays.

\paragraph{Insights and implications.}
Our image embedding visualization reveals that generated images have a different distribution from their prompts in the CLIP embedding space.
For example, the ``movie'' cluster in the prompt embedding has been replaced by the ``portrait'' cluster in the image embedding.
This suggests the semantic representations of prompts and their generated images may not be perfectly aligned.
One hypothesis is that large image generative models face limitations when generating photorealistic human faces~\cite{borjiGeneratedFacesWild2022}, and therefore some images generated with movie-related prompts appear to be closer to art and portrait regions in the embedding space.

\subsection{Stable Diffusion Error Analysis}
\label{sec:analysis:error}
We leverage \tool{} to discover Stable Diffusion generation failure cases and examine potential causes.
To surface poor image generations, we compute CLIP embeddings for all prompts and images in \tool{}.
We then select prompt-image pairs with a large \textcolor{deeporangeIX}{cosine distance ($d$)} between their embeddings.
The cosine distances have a normal distribution~($\mathcal{N}(0.7123, 0.0413^2)$\inlinefig{10}{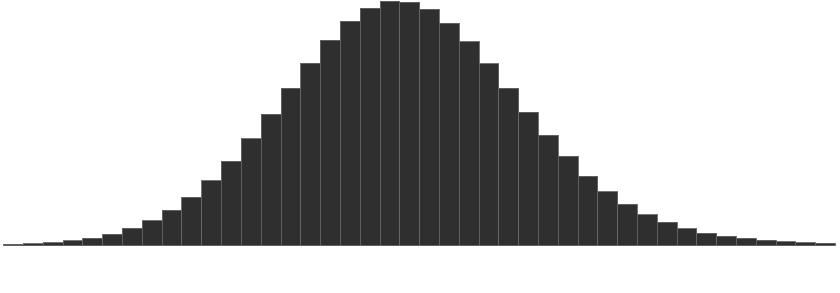}).
In this analysis, we focus on 13,411 ``bad'' prompt-image pairs (1) with a distance that is larger than 4 standard deviations from the mean and (2) the image was not blurred by Stable Diffusion~(\autoref{sec:construction:nsfw}).

\paragraph{Impacts of hyperparameters.}
We conduct a logistic regression test to analyze the relationship between Stable Diffusion hyperparameter values (e.g., \texttt{CFG scale}, \texttt{step}, \texttt{width}, and \texttt{height}) and the likelihood of generating an image that is semantically different from its prompt.
The results reveal that all four hyperparameters are negatively correlated with the likelihood of generating a bad image.
The correlation is statistically significant with a $p$-value of less than 0.0001 for all four variables.
Furthermore, we find the distribution of selected \texttt{sampler} options when generating bad images is significantly different from the overall distribution~($X^2=40873.11$, $p < 0.0001$).

\setlength{\columnsep}{5pt}%
\setlength{\intextsep}{-3pt}%
\begin{wrapfigure}{R}{0.26\textwidth}
  \vspace{0pt}
  \centering
  \includegraphics[width=0.26\textwidth]{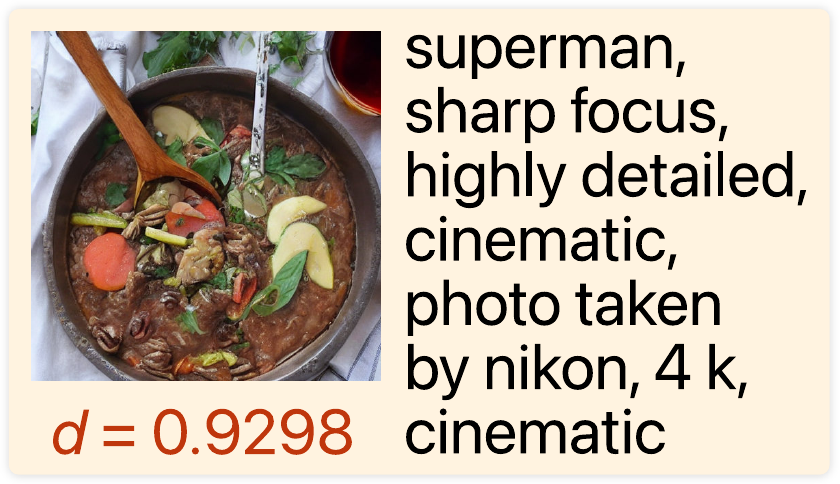}
  \vspace{-10pt}
  \label{fig:error-view-cfg}
\end{wrapfigure}
\texttt{CFG scale} controls how much the generated image looks like the prompt.
We find some users specify negative \texttt{CFG scales} that make images look different from their prompts (large \textcolor{deeporangeIX}{cosine distance $d$}).
In the example shown on the right, a user generates an image using a prompt about ``\texttt{superman}'' with all default hyperparameters values, except for setting \texttt{CFG scale} to \texttt{-1}.
This results in an image featuring a bowl of soup instead of ``\texttt{superman}''.

\setlength{\columnsep}{6pt}
\setlength{\intextsep}{-15pt}
\begin{wrapfigure}{R}{0.26\textwidth}
  \vspace{-4pt}
  \centering
  \includegraphics[width=0.26\textwidth]{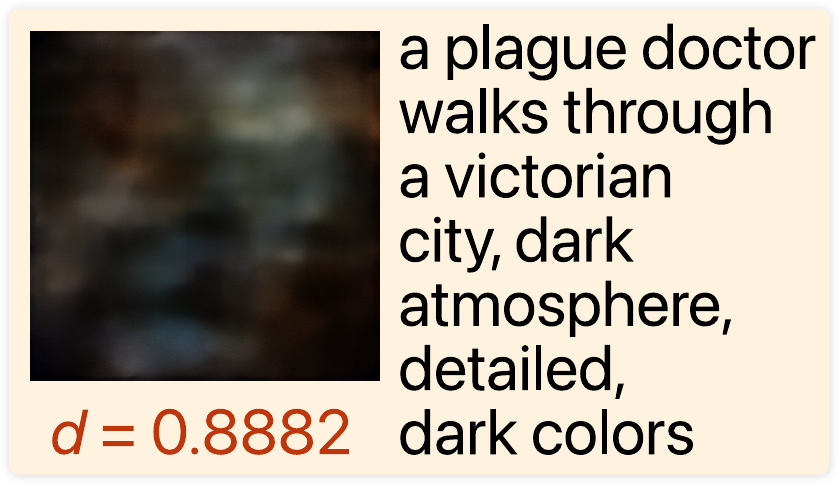}
  \label{fig:error-view-step}
\end{wrapfigure}
A small \texttt{step} could also generate under-developed images that look different from the specified prompts.
As demonstrated in the example on the right, a user generates an image about ``\texttt{plague doctor}'' with all default hyperparameter values, except for setting \texttt{step} to \texttt{2}, which leads to a blurry image.

\setlength{\columnsep}{6pt}%
\setlength{\intextsep}{-3pt}%
\begin{wrapfigure}{R}{0.2\textwidth}
  \vspace{0pt}
  \centering
  \includegraphics[width=0.2\textwidth]{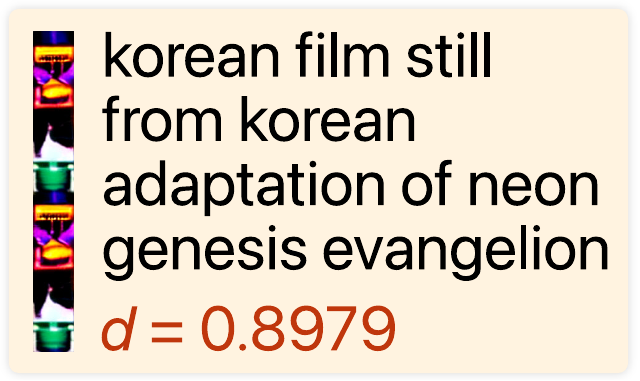}
  \vspace{-10pt}
  \vspace{0pt}
  \label{fig:error-view-size}
\end{wrapfigure}
Stable Diffusion struggles with generating images with a small \texttt{size} or large aspect ratios.
The dissimilar image shown on the right is generated with default hyperparameters except for a \texttt{size} of \texttt{(64,512)}.

\paragraph{Impacts of prompts.}
\setlength{\columnsep}{6pt}%
\setlength{\intextsep}{2pt}%
\begin{wrapfigure}{R}{0.13\textwidth}
  \vspace{-2pt}
  \centering
  \includegraphics[width=0.13\textwidth]{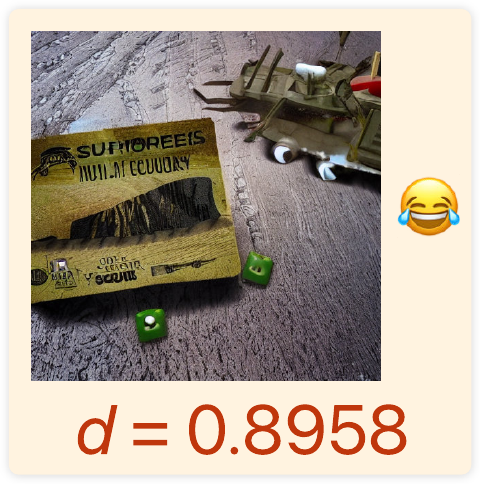}
  \vspace{-20pt}
  \vspace{0pt}
  \label{fig:error-view-prompt}
\end{wrapfigure}
Despite controlling all hyperparameters to be close to default values, we still find 1.1k unique bad image-prompt pairs.
Most of these instances have non-English prompts, very short prompts, or prompts consisting primarily emojis (see an example on the right).
The token lengths of these instances are significantly lower than the overall token length~(one-tailed $t=-23.7203$, $p<0.0001$).
The English prompt frequency among these instances is also significantly lower than the overall frequency~($X^2=1024.56$, $p<0.0001$).
Interestingly, we also find that Stable Diffusion sometimes generates unexpected images even when prompts are meaningful English sentences.
Future researchers can use our error analysis and failure cases to check potentially mislabeled training data.

\paragraph{Implications.}
Our study reveals Stable Diffusion can make mistakes when generating images with certain hyperparameter values or prompt styles.
Negative \texttt{CFG scales}, small \texttt{steps}, or small \texttt{sizes} contributes to generating images dissimilar to prompts.
Short and non-English prompts can also lead to errors.
To improve the quality of future generative models, researchers can expand the training data to cover these edge cases.
There are opportunities for researchers to design user interfaces that can help users understand the impact of different hyperparameters and guide them in choosing values that fit their specific use cases.

\subsection{Potentially Harmful Uses}
\label{sec:analysis:harmful}

To identify potentially malicious uses of Stable Diffusion, we use named entity recognition to analyze prompts.
We find that many prompts include names of influential politicians, such as over 65k images generated with a prompt including ``\texttt{Donald Trump}'' and over 48k images with ``\texttt{Joe Biden}.''
Some prompts portray these politicians in negative lights, ranging from depicting them ``\texttt{as Gollum with hair}'' to ``\texttt{arrested in handcuffs}.''
Additionally, we find female celebrities are frequently used in prompts, with a high frequency after artists and influential politicians.
Some of these prompts are presented in a sexual context that could be considered nonconsensual pornography.

Through keyword search, we discover prompts generating misinformation that could cause harm.
For example, the prompt "\texttt{scientists putting microchips into a vaccine}" may harm public trust in medical institutions by potentially validating conspiracy theories.
Similarly, the prompt "\texttt{Russian soldiers in gas masks found the last surviving ukrainian after a nuclear war to liberate ukraine}" depicts false images of the Russo-Ukrainian War and could lead to new forms of propaganda.
Our findings highlight the crucial need for further research on the broader impacts of large generative models and ways to regulate and mitigate their harms. %
\section{Enabling New Research Directions}
\label{sec:future}

The unprecedented scale and diversity of \tool{} bring new exciting research opportunities to help users generate images more effectively and efficiently, and enable researchers to improve, explain, and safeguard generative models.

\paragraph{Prompt Autocomplete.}
\label{sec:future:autocomplete}

With \tool{}, researchers can develop an autocomplete system to help users construct prompts.
For example, one can use the prompt corpus to train an $n$-gram model to predict likely words following a prompt part.
Alternatively, researchers can use \textit{semantic autocomplete}~\cite{hyvonenSemanticAutocompletion2006} by categorizing prompt keywords into ontological categories such as subject, style, quality, repetition, and magic terms~\cite{oppenlaenderTaxonomyPromptModifiers2022}.
This allows the system to suggest related keywords from unspecified categories, for example suggesting style keyword ``\texttt{depth of field}'' and a magic keyword ``\texttt{award-winning}'' to improve the quality of generated images.
Additionally, researchers can also use \tool{} to study prompt \textit{auto-replace} by distilling effective prompt patterns and creating a ``translation'' model that replaces weaker prompt keywords with more effective ones.

\paragraph{Generation through Search.}
\label{sec:future:search}

As \tool{} contains \dbsize{} images, this dataset might have already included images with a user's desired effects.
Thus, a user can quickly search images in \tool{} instead of running Stable Diffusion, which can be slow and costly.
Lexica~\cite{shameemLexicaBuildingCreative2022}, an AI start-up, provides such a search engine, where users can search Stable Diffusion images by natural language or images.
Researchers can also construct a structured index of images and prompts, such as building a \textit{semantivisual image hierarchy} of images~\cite{liBuildingUsingSemantivisual2010} or a \textit{hierarchical topic model} of prompts~\cite{griffithsHierarchicalTopicModels2003}, to help users easily discover and explore images and prompts with similar styles.

\paragraph{Improving Generative Models.}
\label{sec:future:improve}

With \tool{}, a large and diverse collection of Stable Diffusion usage logs, researchers not only can identify weak points and failure modes of Stable Diffusion but also gain insights into user preferences.
For example, we demonstrate that researchers can use joint text-image embeddings between prompts and images to detect generation misalignments~(\autoref{sec:analysis:error}).
Additionally, \tool{} provides important metadata such as \texttt{username hash} and \texttt{timestamp} for each generated image. By analyzing these metadata fields, researchers can trace the evolution chain of prompts, parameters, and images, which offers valuable insights into how users develop mental models of large generative models and their preferences of generated images.
This understanding can inform future researchers to enhance generative models and design interfaces that facilitate better image-generation experiences.

\paragraph{Explainable Generation.}
\label{sec:future:explain}

As generative models have been gaining immense popularity, there is a call for explainable creativity~\cite{llanoExplainableComputationalCreativity2022}.
Many explanation techniques use input permutation that computes feature attribution scores by running a model on slightly-modified input values~\cite{lundbergUnifiedApproachInterpreting2017}.
\tool{} contains \dbsize{} prompt-image pairs including similar prompts with minor differences, such as ``\texttt{a happy dog}'' and ``\texttt{a sad dog}'', allowing researchers to investigate how individual keywords affect the generation process.

\paragraph{Deepfake Detection.}
\label{sec:future:fake}

Breakthroughs in generative models raise concerns about deepfakes---fake images of real individuals for unethical purposes~\cite{wiggersDeepfakesAllUncensored2022}.
\tool{} is valuable for detecting deepfakes, as it contains a large-scale collection of model-generated images and their metadata.
Researchers can use this collection to train ML models to identify synthetic artifacts and train classifiers that classify synthetic images from real images~\cite{mirskyCreationDetectionDeepfakes2022}.

\section{Related Work}

\paragraph{Text-to-text Prompting.}
Researchers have been studying prompt engineering for text-to-text generation~\cite[e.g.,][]{liuPretrainPromptPredict2022,luFantasticallyOrderedPrompts2022,rubinLearningRetrievePrompts2022}.
To facilitate this line of research, researchers develop PromptSource~\cite{bachPromptSourceIntegratedDevelopment2022}, a dataset of 2k text prompts along with a framework to create and share prompts.
In contrast, our work focuses on text-to-image prompting, and \tool{} has an unprecedented scale of \dbsize{} real prompt-image pairs.

\paragraph{Text-to-image Prompting.}
There is a growing interest in text-to-image prompt engineering research from NLP, Computer Vision, and HCI communities~\cite[e.g.,][]{qiaoInitialImagesUsing2022, pavlichenkoBestPromptsTexttoImage2022}.
For example, \citet{oppenlaenderTaxonomyPromptModifiers2022} identifies six types of prompt modifiers through an ethnographic study, and \citet{liuDesignGuidelinesPrompt2022} proposes design guidelines for text-to-image prompt engineering by experimenting with 1,296 prompts.
Closest in spirit to \tool{} is Lexica~\cite{shameemLexicaBuildingCreative2022} which allows users to search over 5 million Stable Diffusion images with their prompts, but it does not release its internal database.
In comparison, \tool{} is open-source and publicly available to everyone. %
\section{Conclusion}

We present \tool{}, the first large-scale text-to-image prompt dataset, containing \dbsize{} images with their prompts and hyperparameters collected from the Stable Diffusion discord server.
We release the dataset with a CC0 1.0 license and open source all collection and analysis code, broadening the public's access to cutting-edge AI technologies.
We discuss findings on prompt and image patterns.
We hope our work will serve as a cornerstone for the future development of large generative modes and tools that help users use these modes.

\section{Limitations}
\label{sec:limitations}

We discuss four limitations of our work: the inclusion of unsafe content, potential biases in data sources, a limited measure of image quality and generalizability to different generative models.

\begin{itemize}[topsep=3pt, itemsep=2pt, parsep=0pt, leftmargin=10pt]
  \item \textbf{Inclusion of unsafe images and prompts.}
  We collect images and their prompts from the Stable Diffusion Discord server~(\autoref{sec:construction}).
  The Discord server has rules against users generating or sharing harmful or NSFW (not suitable for work, such as sexual and violent content) images.
  The Stable Diffusion model used in the server also has an NSFW filter that blurs the generated images if it detects NSFW content.
  However, we observe that \tool{} includes some NSFW images that were not detected by the NSFW filter or removed by the server moderators.
  To mitigate the potential harm, we compute and share the likelihood of an image or a prompt containing unsafe content using the state-of-the-art NSFW detectors~(\autoref{sec:construction:nsfw}).
  In addition, we provide a Google Form on the \tool{} website where users can report harmful or inappropriate images and prompts.
  We will closely monitor this form and remove reported images and prompts from \tool{}.

  \item \textbf{Potential biases of the data source.}
  The \dbsize{} images in \tool{} have diverse styles and categories.
  However, Discord can be a biased data source.
  Our images come from channels where early users could use a bot to use Stable Diffusion before release.
  As these users had started using Stable Diffusion before the model was public, we hypothesize that they are AI art enthusiasts and are likely to have experience with other text-to-image generative models.
  Therefore, the prompting style in \tool{} might not represent novice users.
  Similarly, the prompts in \tool{} might not generalize to domains that require specific knowledge, such as medical images~\cite{chambonAdaptingPretrainedVisionLanguage2022}.\looseness=-1

  \item \textbf{Limited measure of image quality.}
  We use joint text-image CLIP embeddings between prompts and images to detect generation misalignments~(\autoref{sec:analysis:error}).
  While the CLIP embedding distance can indicate the degree of alignment between the prompts and generated images, it does not provide a measure of the overall image quality.
  When constructing our dataset, we have considered including image properties such as entropy, variance, and the most common colors to help users gauge image qualities. However, these metrics do not provide a good measure of the overall image quality as well.
  To better measure image quality, future researchers can recruit annotators to rate images in \tool{}.

  \item \textbf{Generalizability.}
  Previous research has shown a prompt that works well on one generative model might not give the optimal result when used in other models~\cite{borjiGeneratedFacesWild2022}.
  Therefore, different models can need users to write different prompts.
  For example, many Stable Diffusion prompts use commas to separate keywords, while this pattern is less seen in prompts for DALL-E 2~\cite{rameshHierarchicalTextConditionalImage2022} or Midjourney~\cite{holzMidjourneyExploringNew2022}.
  Thus, we caution researchers that some research findings from \tool{} might not be generalizable to other text-to-image generative models.

\end{itemize}

\section{Ethics Statement}
\label{sec:ethics}

In this section, we discuss two main ethical considerations of \tool{}.

\begin{itemize}[topsep=3pt, itemsep=2pt, parsep=0pt, leftmargin=10pt]
  \item \textbf{Copyright.}
  By using the Stable Diffusion Discord server, all users agree to the entirety of CC0 1.0 Universal Public Domain Dedication.
  This includes waiving any intellectual property rights related to any content shared on the server~\cite{stabilityaiStableDiffusionDream2022}.
  All prompts and images in the Discord server are considered to be public domain and can be used by anyone for any purpose.
  Also, we release \tool{} under the CC0 1.0 license~(\autoref{sec:construction:distribution}).

  \item \textbf{Privacy.}
  While it is possible that some prompts may contain sensitive information, this is not common because the Stable Diffusion Discord has strict rules against writing personal information in the prompts and has moderators in place to remove violative messages.
  To further protect user privacy, we have anonymized the usernames of all users in our dataset~(\autoref{sec:construction:organize}).
  Users also have the option to remove their prompts and images from our dataset through an online form~(\autoref{sec:construction:distribution}).
\end{itemize}

\noindent We provide a thorough discussion on the limitations and broader impacts of \tool{} in its Data Sheet~\cite{gebruDatasheetsDatasets2020}~(\autoref{sec:datasheet}). 

\section*{Acknowledgements}
We thank Stability AI for releasing Stable Diffusion and hosting the Stable Diffusion Discord server.
We especially appreciate the Stable Diffusion Discord moderators and users for creating an open and friendly online community that makes our work possible.
We also extend our appreciation to Hugging Face for hosting our dataset.
Lastly, we would like to acknowledge the anonymous reviewers for their valuable feedback and insightful comments that helped improve our paper.
This work was supported in part by a J.P. Morgan PhD Fellowship, NSF grants IIS-1563816, DARPA GARD, gifts from Cisco, Bosch, and NVIDIA.
Use, duplication, or disclosure is subject to the restrictions as stated in Agreement number HR00112030001 between the Government and the Performer.

\balance
\bibliographystyle{acl_natbib}
\bibliography{23-diffusiondb}

\clearpage
\appendix

\section{Data Sheet for \tool{}}
\label{sec:datasheet}

\noindent\fbox{\begin{minipage}{\linewidth}
  \vspace{3pt}
  \centering
  \large{\textbf{\dscolor{Motivation}}}
  \vspace{3pt}
\end{minipage}}
\\

\noindent
\dscolor{\textbf{For what purpose was the dataset created?} Was there a specific task in mind? Was there a specific gap that needed to be filled? Please provide a description.}
\\
The \tool{} project was inspired by important needs in research focused on diffusion models and prompt engineering. As large text-to-image models are relatively new, there is a pressing need to understand how these models work, how to write effective prompts, and how to design tools to help users generate images. To tackle these critical challenges, we present \tool{}, the first large-scale prompt dataset with \dbsize{} real prompt-image pairs.
\\ \\
\dscolor{\textbf{Who created the dataset (e.g., which team, research group) and on
behalf of which entity (e.g., company, institution, organization)?}} \\
The dataset was created by Zijie J. Wang, Evan Montoya, David Munechika, Haoyang Yang, Benjamin Hoover, and Duen Horng Chau at the Georgia Institute of Technology.
\\ \\
\dscolor{\textbf{Who funded the creation of the dataset?} If there is an associated grant, please provide the name of the grantor and the grant name and number.} \\
Funded in part by J.P. Morgan PhD Fellowship, NSF grants IIS-1563816, DARPA GARD, and gifts from Cisco, Bosch, and NVIDIA.
\\ \\
\dscolor{\textbf{Any other comments?}} \\
None.\\

\noindent\fbox{\begin{minipage}{\linewidth}
  \vspace{3pt}
  \centering
  \large{\textbf{\dscolor{Composition}}}
  \vspace{3pt}
\end{minipage}}
\\

\noindent
\dscolor{\textbf{What do the instances that comprise the dataset represent (e.g., documents, photos, people, countries)?} Are there multiple types of instances (e.g., movies, users, and ratings; people and interactions between them; nodes and edges)? Please provide a description.} \\
Each instance consists of an image generated by the Stable Diffusion model and the prompt as well as parameters that were input into the model to generate the image.
The input parameters include \texttt{seed}, \texttt{CFG scale}, \texttt{sampler}, \texttt{width}, \texttt{height}, \texttt{username hash}, \texttt{timestamp}, \texttt{image NSFW score} and \texttt{prompt NSFW score}.
\\ \\
\dscolor{\textbf{How many instances are there in total (of each type, if appropriate)?}} \\
There are \dbsize{} instances in total.
\\ \\
\dscolor{\textbf{Does the dataset contain all possible instances or is it a sample (not necessarily random) of instances from a larger set?} If the dataset is
a sample, then what is the larger set? Is the sample representative of the
larger set (e.g., geographic coverage)? If so, please describe how this
representativeness was validated/verified. If it is not representative of the larger set, please describe why not (e.g., to cover a more diverse range of instances, because instances were withheld or unavailable).} \\
The dataset is a sample of instances. It represents a sample of images from the Stable Diffusion discord server. No tests were run to determine representativeness.
\\ \\
\dscolor{\textbf{What data does each instance consist of?} “Raw” data (e.g., unprocessed text or images)or features? In either case, please provide a description.} \\
Each instance consists of the image generated by the Stable Diffusion model (with a unique id), along with the prompt used to generate the image and the model parameters as a JSON file.
\\ \\
\dscolor{\textbf{Is there a label or target associated with each instance?} If so, please provide a description.} \\
The labels associated with each image are the prompt and other input parameters.
\\ \\
\dscolor{\textbf{Is any information missing from individual instances?} If so, please provide a description, explaining why this information is missing (e.g., because it was unavailable). This does not include intentionally removed information, but might include, e.g., redacted text.} \\
Everything is included. No data is missing.
\\ \\
\dscolor{\textbf{Are relationships between individual instances made explicit (e.g.,
users’ movie ratings, social network links)?} If so, please describe
how these relationships are made explicit.} \\
Not applicable.
\\ \\
\dscolor{\textbf{Are there recommended data splits (e.g., training, development/validation, testing)?} If so, please provide a description of these splits, explaining the rationale behind them.} \\
No. This dataset is not for ML model benchmarking. Researchers can use any subsets of it.
\\ \\
\dscolor{\textbf{Are there any errors, sources of noise, or redundancies in the
dataset?} If so, please provide a description.} \\
No. All images and prompts are extracted as is from the Discord chat log.
\\ \\
\dscolor{\textbf{Is the dataset self-contained, or does it link to or otherwise rely on external resources (e.g., websites, tweets, other datasets)?}} \\
The dataset is entirely self-contained.
\\ \\
\dscolor{\textbf{Does the dataset contain data that might be considered confidential
(e.g., data that is protected by legal privilege or by doctor–patient
confidentiality, data that includes the content of individuals’ nonpublic communications)?} If so, please provide a description.
Unknown to the authors of the datasheet.} \\
It is possible that some prompts contain sensitive information. However, it would be rare, as the Stable Diffusion Discord has rules against writing personal information in the prompts, and there are moderators removing messages that violate the Discord rules.
\\ \\
\dscolor{\textbf{Does the dataset contain data that, if viewed directly, might be offensive, insulting, threatening, or might otherwise cause anxiety?} If so, please describe why.} \\
We collect images and their prompts from the Stable Diffusion discord server. Even though the discord server has rules against users sharing any NSFW (not suitable for work, such as sexual and violent content) and illegal images, \tool{} still contains some NSFW images and prompts that were not removed by the server moderators.
\\ \\
\dscolor{\textbf{Does the dataset identify any subpopulations (e.g., by age, gender)?} If so, please describe how these subpopulations are identified and provide a description of their respective distributions within the dataset.} \\
No.
\\ \\
\dscolor{\textbf{Is it possible to identify individuals (i.e., one or more natural persons), either directly or indirectly (i.e., in combination with other
data) from the dataset?} If so, please describe how.} \\
No.
\\
\dscolor{\textbf{Any other comments?}} \\
None. \\

\noindent\fbox{\begin{minipage}{\linewidth}
  \vspace{3pt}
  \centering
  \large{\textbf{\dscolor{Collection}}}
  \vspace{3pt}
\end{minipage}}
\\

\noindent
\dscolor{\textbf{How was the data associated with each instance acquired?} Was
the data directly observable (e.g., raw text, movie ratings), reported by
subjects (e.g., survey responses), or indirectly inferred/derived from other
data (e.g., part-of-speech tags, model-based guesses for age or language)?
If the data was reported by subjects or indirectly inferred/derived from
other data, was the data validated/verified? If so, please describe how.} \\
The data was directly observed from the Stable Diffusion Discord Channel. It was gathered from channels where users can generate images by interacting with a bot, which consisted of messages of user generated images and the prompts used to generate those images.
\\ \\
\dscolor{\textbf{What mechanisms or procedures were used to collect the data
(e.g., hardware apparatuses or sensors, manual human curation,
software programs, software APIs)?} How were these mechanisms or
procedures validated?} \\
The data was gathered using a DiscordChatExporter~\cite{holubDiscordChatExporterExportsDiscord2017}, which collected images and chat messages from each channel specified.
We then extracted and linked prompts to images using Beautiful Soup~\cite{richardsonBeautifulSoupDocumentation2007}.
Random images and prompts were selected and manually verified to validate the prompt-image mapping.
\\ \\
\dscolor{\textbf{If the dataset is a sample from a larger set, what was the sampling
strategy (e.g., deterministic, probabilistic with specific sampling
probabilities)?}} \\
\tool{} does not sample from a larger set.
However, \tool{}-2M is a sample from a larger set.
For certain messages, there would exist a collage of $n$ images (e.g., $n$ = 2, 4, 9) with identical prompts consolidated into a single image.
These images were split and a single image would be randomly selected to include in \tool{}-2M from $n$ images with equal probability of any image being selected.
This saved space and prioritized unique prompts.
\\ \\
\dscolor{\textbf{Who was involved in the data collection process (e.g., students,
crowdworkers, contractors) and how were they compensated (e.g.,
how much were crowdworkers paid)?}} \\
Students conducted the data collection process and were compensated with stipend or course credits.
\\ \\
\dscolor{\textbf{Over what timeframe was the data collected? Does this timeframe
match the creation timeframe of the data associated with the instances
(e.g., recent crawl of old news articles)?} If not, please describe the timeframe in which the data associated with the instances was created.} \\
All messages were generated in August 2022 and messages were collected between October 18th and 24th 2022.
\tool{} includes the generation timestamps of all images.
\\ \\
\dscolor{\textbf{Were any ethical review processes conducted (e.g., by an institutional review board)?} If so, please provide a description of these review processes, including the outcomes, as well as a link or other access point to any supporting documentation.} \\
There were no ethical review processes conducted.
\\ \\
\dscolor{\textbf{Did you collect the data from the individuals in question directly,
or obtain it via third parties or other sources (e.g., websites)?}} \\
The data was directly obtained from individual messages in the Discord server.
\\ \\
\dscolor{\textbf{Were the individuals in question notified about the data collection?} If so, please describe (or show with screenshots or other information) how notice was provided, and provide a link or other access point to, or otherwise reproduce, the exact language of the notification itself.}\\
Users of the channel were not notified about this specific gathering of data but agree to forfeit any intellectual property rights claims by using Stable Diffusion.
In addition, users are instructed that the images are public domain and can be used by anyone for any purpose. The exact language is as follows~\cite{stabilityaiStableDiffusionDream2022}:
\begin{displayquote}
Note, that while users have forfeited copyright (and any/all intellectual property right claims) on these images, they are still public domain and can be used by anyone for any purpose, including by the user. Feel free to use images from DreamStudio Beta and the Stable Diffusion beta Discord service for anything, including commercial purposes.
\end{displayquote}

\noindent{}\dscolor{\textbf{Did the individuals in question consent to the collection and use
of their data?} If so, please describe (or show with screenshots or other
information) how consent was requested and provided, and provide a
link or other access point to, or otherwise reproduce, the exact language
to which the individuals consented.} \\
By using the server and tools, users consented to the regulations posed by Stability AI LTD, the company that both made Stable Diffusion and runs the Discord server. This implies consent by using the tool. The exact wording is as follows:
\begin{displayquote}
  By your use of DreamStudio Beta and the Stable Diffusion, you hereby agree to forfeit all intellectual property rights claims, worldwide, and regardless of legal jurisdiction or intellectual property law applicable therein, including forfeiture of any/all copyright claim(s), to the Content you provide or receive through your use of DreamStudio Beta and the Stable Diffusion beta Discord service.
\end{displayquote}

\noindent{}This message is contained in the rules and terms of service section of the Stable Diffusion Discord~\cite{stabilityaiStableDiffusionDiscord2022,stabilityaiStableDiffusionDream2022}.
In conjunction with the previous statement about images being public domain (CC0 1.0 license), it is established that the images made by using Stable Diffusion can be used for other purposes.
\\ \\
\dscolor{\textbf{If consent was obtained, were the consenting individuals provided with a mechanism to revoke their consent in the future or for certain uses?} If so, please provide a description, as well as a link or other access point to the mechanism (if appropriate).} \\
Users will have the option to report harmful content or withdraw images they created through a Google Form listed on the \tool{} website: \textbf{\url{https://github.com/poloclub/diffusiondb}}.
\\ \\
\dscolor{\textbf{Has an analysis of the potential impact of the dataset and its use
on data subjects (e.g., a data protection impact analysis) been conducted?} If so, please provide a description of this analysis, including
the outcomes, as well as a link or other access point to any supporting
documentation.} \\
No analysis has been conducted.
\\ \\
\dscolor{\textbf{Any other comments?}} \\
None.
\\ \\
\noindent\fbox{\begin{minipage}{\linewidth}
  \vspace{3pt}
  \centering
  \large{\textbf{\dscolor{Preprocessing}}}
  \vspace{3pt}
\end{minipage}}
\\

\noindent{}\dscolor{\textbf{Was any preprocessing/cleaning/labeling of the data done (e.g.,
discretization or bucketing, tokenization, part-of-speech tagging,
SIFT feature extraction, removal of instances, processing of missing values)?} If so, please provide a description. If not, you may skip the
remaining questions in this section.} \\
The Discord chat logs include collage images, where each collage contains a grid of images that share the same prompt but have different seeds.
We use Pillow~\cite{clarkPillowPythonImaging2015} to split a collage into individual images.
For \tool{}, we include all split images.
However, for \tool{}-2M, we only include one randomly selected split image to save space and prioritize unique prompts.
\\ \\
\dscolor{\textbf{Was the “raw” data saved in addition to the preprocessed/cleaned/labeled
data (e.g., to support unanticipated future uses)?} If so, please provide a link or other access point to the “raw” data.} \\
Raw data was not saved.
\\ \\
\dscolor{\textbf{Is the software that was used to preprocess/clean/label the data
available?} If so, please provide a link or other access point.} \\
All our data collection and preprocessing code is available at: \textbf{\url{https://github.com/poloclub/diffusiondb}}.
\\ \\
\dscolor{\textbf{Any other comments?}}\\
None.
\\

\noindent\fbox{\begin{minipage}{\linewidth}
  \vspace{3pt}
  \centering
  \large{\textbf{\dscolor{Uses}}}
  \vspace{3pt}
\end{minipage}}
\\

\noindent
\dscolor{\textbf{Has the dataset been used for any tasks already?} If so, please provide a description.}\\
No.
\\\\
\dscolor{\textbf{Is there a repository that links to any or all papers or systems that use the dataset?} If so, please provide a link or other access point.}\\
No.
\\\\
\dscolor{\textbf{What (other) tasks could the dataset be used for?}}\\
This dataset can be used for (1) prompt autocomplete, (2) generating images through search, (3) detecting deepfake, (4) debugging image generation, (5) explaining image generation, and more.
\\\\
\dscolor{\textbf{Is there anything about the composition of the dataset or the way it was collected and preprocessed/cleaned/labeled that might impact future uses? } For example, is there anything that a dataset consumer might need to know to avoid uses that could result in unfair treatment of individuals or groups (e.g., stereotyping, quality of service issues) or other risks or harms (e.g., legal risks, financial harms)? If so, please provide a description. Is there anything a dataset consumer could do to mitigate these risks or harms?}\\
There is minimal risk for harm: the data were already public.
Personally identifiable data (e.g., discord usernames) were removed during the collection/preprocessing phases.
\\\\
\dscolor{\textbf{Are there tasks for which the dataset should not be used?} If so, please provide a description.}\\
All tasks that utilize this dataset should follow the licensing policies and the regulations~\cite{stabilityaiStableDiffusionDream2022} posed by Stability AI, the company that both made Stable Diffusion and runs the official Discord server.
\\\\
\dscolor{\textbf{Any other comments?}}\\ \
None.
\\

\noindent\fbox{\begin{minipage}{\linewidth}
  \vspace{3pt}
  \centering
  \large{\textbf{\dscolor{Distribution}}}
  \vspace{3pt}
\end{minipage}}
\\

\noindent
\dscolor{\textbf{Will the dataset be distributed to third parties outside of the entity (e.g., company, institution, organization) on behalf of which the dataset was created?} If so, please provide a description.}\\
Yes, the dataset is publicly available on the internet.
\\ \\
\dscolor{\textbf{How will the dataset will be distributed (e.g., tarball on website, API, GitHub)?} Does the dataset have a digital object identifier (DOI)?}\\
The dataset is distributed on the project website: \textbf{\url{https://poloclub.github.io/diffusiondb}}.
The dataset shares the same DOI as this paper.
\\ \\
\dscolor{\textbf{When will the dataset be distributed?}}\\
The dataset is released on October 25th, 2022.
\\\\
\dscolor{\textbf{Will the dataset be distributed under a copyright or other intellectual property (IP) license, and/or under applicable terms of use (ToU)?} If so, please describe this license and/or ToU, and provide a link or other access point to, or otherwise reproduce, any relevant licensing terms or ToU, as well as any fees associated with these restrictions.}\\
All images generated by stable diffusion discord services are under the \href{https://creativecommons.org/publicdomain/zero/1.0}{CC0 1.0 License}, and therefore so are images in this dataset.
In addition, the distribution of the dataset is under the Terms of Use~\cite{stabilityaiStableDiffusionDream2022} posed by
Stability AI, the company that both made Stable Diffusion and runs the official Discord server.
\\\\
\dscolor{\textbf{Have any third parties imposed IP-based or other restrictions on the data associated with the instances?} If so, please describe these restrictions, and provide a link or other access point to, or otherwise reproduce, any relevant licensing terms, as well as any fees associated with these restrictions.}\\
All images in this dataset have a \href{https://creativecommons.org/publicdomain/zero/1.0}{CC0 1.0 License} and follows the Stability AI's Terms of Use~\cite{stabilityaiStableDiffusionDream2022}.
\\\\
\dscolor{\textbf{Do any export controls or other regulatory restrictions apply to the dataset or to individual instances?} If so, please describe these restrictions, and provide a link or other access point to, or otherwise reproduce, any supporting documentation.}\\
No.
\\\\
\dscolor{\textbf{Any other comments?}}\\
None.
\\

\noindent\fbox{\begin{minipage}{\linewidth}
  \vspace{3pt}
  \centering
  \large{\textbf{\dscolor{Maintenance}}}
  \vspace{3pt}
\end{minipage}}
\\\\
\dscolor{\textbf{Who will be supporting/hosting/maintaining the dataset?}}\\
The authors of this paper will be supporting and maintaining the dataset.
\\\\
\dscolor{\textbf{How can the owner/curator/manager of the dataset be contacted (e.g., email address)?}}\\
The contact information of the curators of the dataset is listed on the project website: \textbf{\url{https://poloclub.github.io/diffusiondb}}.
\\\\
\dscolor{\textbf{Is there an erratum?} If so, please provide a link or other access point.}\\
There is no erratum for our initial release.
Errata will be documented in future releases on the dataset website.
\\\\
\dscolor{\textbf{Will the dataset be updated (e.g., to correct labeling errors, add new instances, delete instances)?} If so, please describe how often, by whom, and how updates will be communicated to dataset consumers (e.g., mailing list, GitHub)?}\\
Yes, we will monitor the Google Form where users can report harmful images and creators can remove their images.
We will update the dataset bimonthly.
Updates will be posted on the project website \textbf{\url{https://poloclub.github.io/diffusiondb}}.
\\\\
\dscolor{\textbf{If the dataset relates to people, are there applicable limits on the retention of the data associated with the instances (e.g., were the individuals in question told that their data would be retained for a fixed period of time and then deleted)?} If so, please describe these limits and explain how they will be enforced.}\\
People can use a Google Form linked on the project website to remove specific instances from \tool{}.
\\\\
\dscolor{\textbf{Will older versions of the dataset continue to be supported/hosted/maintained? }If so, please describe how. If not, please describe how its obsolescence will be communicated to dataset consumers.}\\
We will continue to support older versions of the dataset.
\\\\
\dscolor{\textbf{If others want to extend/augment/build on/contribute to the dataset, is there a mechanism for them to do so?} If so, please provide a description. Will these contributions be validated/verified? If so, please describe how. If not, why not? Is there a process for communicating/distributing these contributions to dataset consumers? If so, please provide a description.}\\
Anyone can extend/augment/build on/contribute to \tool{}.
Potential collaborators can contact the dataset authors.
\\\\
\textbf{Any other comments?}\\
None. 
\end{document}